\documentclass{article}

\usepackage{microtype}
\usepackage{graphicx}
\usepackage{subcaption}
\usepackage{booktabs} % for professional tables

\usepackage{hyperref}

\usepackage[preprint]{icml2026}

\usepackage{amsmath}
\usepackage{amssymb}
\usepackage{mathtools}
\usepackage{amsthm}

\usepackage{amsfonts}
\usepackage{subcaption}
\usepackage{booktabs}
\usepackage{makecell}
\usepackage{adjustbox}
\usepackage{multirow}
\usepackage{diagbox}
\usepackage{algorithm}
\usepackage{algorithmic}
\usepackage{wrapfig}
\usepackage{bm}
\usepackage{enumitem}
\usepackage{tabularx}
\usepackage{xcolor}
\usepackage{titlesec}

\usepackage[capitalize,noabbrev]{cleveref}

\theoremstyle{plain}
\newtheorem{theorem}{Theorem}[section]

\newtheorem{lemma}[theorem]{Lemma}

\newtheorem{definition}[theorem]{Definition}

\theoremstyle{remark}

\usepackage[textsize=tiny]{todonotes}

\icmltitlerunning{Reflex: Reinforcement Learning with Reflection Symmetry
Exploitation in State-Based Continuous Control}

\begin{document}

\twocolumn[
  \icmltitle{Reflex: Reinforcement Learning with Reflection Symmetry \\
Exploitation in State-Based Continuous Control}



  \icmlsetsymbol{equal}{*}

  \begin{icmlauthorlist}
    \icmlauthor{Shuai Zhen}{yyy}
    \icmlauthor{Yifan Zhang}{yyy}
    \icmlauthor{Yuling Wang}{comp}
    \icmlauthor{Yanhua Yu}{yyy}
  \end{icmlauthorlist}

  \icmlaffiliation{yyy}{Beijing University of Posts and Telecommunications}
  \icmlaffiliation{comp}{Hangzhou Dianzi University}
%   \icmlaffiliation{sch}{School of ZZZ, Institute of WWW, Location, Country}

  \icmlcorrespondingauthor{Shuai Zhen}{zhenshuai@bupt.edu.cn}
%   \icmlcorrespondingauthor{Firstname2 Lastname2}{first2.last2@www.uk}

  \icmlkeywords{Machine Learning, ICML}

  \vskip 0.3in
]



\printAffiliationsAndNotice{}  % no special notice (required even if empty)

\begin{abstract}
Reinforcement learning has long struggled with poor sample efficiency. One promising approach to mitigate this problem is leveraging group-invariant Markov Decision Processes ($G$-invariant MDPs). Existing works in this direction have primarily focused on image-based RL and rotational symmetry such as $\mathrm{SO(2)}$, leaving state-based RL and reflection symmetry largely underexplored. 
In this work, we focus on state-based continuous control tasks and exploit reflection symmetry by introducing \textbf{Reflex}, a paradigm that seamlessly integrates with both on-policy and off-policy RL algorithms. We formalize two types of reflection—\textit{axial reflection} and \textit{bilateral reflection}, and characterize their corresponding transformations.
Building on a theoretical analysis of symmetry-preserving optimal value functions and policies, Reflex integrates reflection symmetry into policy learning through principled symmetry regularization mechanisms.
We integrate Reflex with PPO and SAC, and evaluate it on a suite of OpenAI Gym and DeepMind Control benchmarks, demonstrating superior performance over standard baselines while improving sample efficiency. Our code is available at \url{https://github.com/TonyStark042/Reflex}.

\end{abstract}

\section{Introduction}

While reinforcement learning (RL) has shown remarkable success in solving continuous control tasks~\cite{schulman2017ppo,haarnoja2018sac,fujimoto2018td3,lillicrap2015ddpg}, improving its sample efficiency remains a core challenge, as effective policy learning often requires extensive environment interaction. 
To address this critical bottleneck, a growing body of work has been proposed with the agent observing images as the input representations.
These methods largely build on the idea by applying structured or stochastic augmentations to the observed images during training. Like RAD~\cite{laskin2020RAD} augments image observations with simple transformations such as random cropping, translation and color jittering without modifying underlying RL objective, and other methods like Drq~\cite{yarats2021DrQ,yarats2021DrQ2} 
and CURL~\cite{laskin2020curl} applied similar augmentation but with auxiliary objectives.

In contrast to image-based RL, state-based RL operates directly on structured vector observations encoding physical quantities rather than high-dimensional pixel inputs, thereby reducing learning complexity and allowing agents to focus on task-relevant dynamics.
However, data augmentation strategies that are effective in image-based RL do not readily transfer to state-based continuous control.
In state-based settings, symmetries are often implicit and task-specific, making it difficult to define valid transformation rules.
Moreover, na\"ively applying perturbations on observations, such as Gaussian noise or random feature scaling~\cite{laskin2020RAD}, may violate physical consistency by inducing state perturbations that are inconsistent with the true environment transitions and rewards.This mismatch can lead to limited performance gains or even performance degradation in practice~\cite{luo2024euclidean}.

Moreover, existing symmetry-aware augmentation methods in continuous control have largely focused on rotational and other symmetries, and are predominantly studied in off-policy RL settings~\cite{corrado2024understanding,wang2022SO2,hu2024revisiting}, while reflection symmetry and on-policy RL have received comparatively less attention.
Reflection symmetry is prevalent and practically important in many continuous control tasks, particularly those involving coordinated left-right behaviors such as gait locomotion, yet principled mechanisms for exploiting such symmetries remain limited.
At the same time, although on-policy RL algorithms generally suffer from lower sample efficiency due to limited data reuse, they offer notable advantages including reduced memory consumption and lower computational overhead by performing policy updates less frequently. Following the above observations, a natural question arises: 

\emph{Can we design a method that improves sample efficiency in both on-policy and off-policy RL for state-based continuous control by systematically exploiting reflection symmetry?}

To answer this question, in this paper, we present \textbf{Reflex} (Reinforcement Learning with \underline{Refl}ection Symmetry \underline{Ex}ploitation), a general paradigm for state-based, model-free, and online RL in environments with reflection symmetry. Specifically, we formalize two types of reflection: \textit{axial reflection} and \textit{bilateral reflection}, and define their corresponding transformations on the state and action spaces. Based on these mappings, we introduce a symmetry consistency regularization that encourages the policy to learn from existing experience under mirrored states, thereby improving sample efficiency. Unlike most prior works that focus primarily on off-policy methods in continuous control, we study the application of Reflex to both on-policy and off-policy RL, using PPO~\cite{schulman2017ppo} and SAC~\cite{haarnoja2018sac} as representative algorithms, and design tailored regularization strategies for each.

We summarize our main contributions as follows:
\begin{itemize}[leftmargin=10pt, itemsep=0pt, topsep=2pt]
    \item We propose Reflex, which provides a reflection symmetry-aware learning paradigm that enhances both on-policy and off-policy RL by explicitly incorporating reflection invariance. In particular, we define two reflection types, \textit{axial} and \textit{bilateral}, and analyze their structural properties and applicability across state-based continuous control.
    \item We theoretically prove the symmetry of optimal policies and introduce a symmetry regularization term that explicitly encourages the policy to be equivalent under reflection transformations, further enhancing the robustness and final performance of the learned policies.
    \item We show that Reflex not only significantly improves sample efficiency but also consistently enhances the final expected return across multiple environments in the DeepMind control benchmark as well as the OpenAI Gym benchmark, demonstrating its effectiveness in both accelerating learning and improving overall performance.

\end{itemize}

\section{Related Work}
\paragraph{Symmetry Exploitation in RL.}
Symmetry provides a powerful inductive bias in RL, enabling agents to share experience across equivalent states. Theoretically, symmetries can be viewed as a special case of MDP homomorphisms, where structured transformations on state–action pairs preserve environment dynamics and rewards~\citep{ravindran2004approximate}. This implies that optimal value functions and policies remain invariant under such transformations~\citep{ravindran2001symmetries, zinkevich2001symmetry}. Formally, these transformations can be modeled by a group \( G \), leading to the notion of \( G \)-invariant MDPs, whose transition and reward functions are invariant under the action of \( g \in G \)~\citep{wang2022SO2}.
With the rise of deep RL, \citet{mahajan2017symmetry} were among the first to apply symmetry-based regularization to DQN, introducing symmetric data augmentation and a symmetry-regularized $Q$-function loss in discrete environments. \citet{abdolhosseini2019Onlearning} extended this idea to continuous locomotion tasks by combining architectural constraints, data duplication, and loss-based regularization to promote symmetric motion. Subsequent studies~\citep{su2024leveraging, ordonez2025morphological, abreu2025addressing} have further demonstrated the potential of leveraging symmetry in policy optimization. However, most of these approaches are limited to discrete settings or image-based RL, and their benefits in high-dimensional state-based continuous control tasks remain inconsistent.
Beyond algorithmic regularization, another line of work incorporates symmetry directly into network architectures through equivariance. Building on the success of group-equivariant neural networks~\citep{cohen2016group, finzi2021practical, gerken2023geometric}, several studies have proposed equivariant policy or value networks for image-based RL~\citep{wang2022SO2, mondal2020group, van2020mdp, nguyen2023equivariant, tangri2024equivariant}.
Although equivariant networks provide appealing theoretical guarantees, they generally introduce substantial computational overhead and require carefully designed architectures tailored to specific symmetry groups.

\paragraph{Data Augmentation in RL.}
Beyond symmetry-specific approaches, a number of works have explored general data augmentation techniques in RL. 
Most studies in this direction have focused on image-based RL~\citep{dmitry2018qt, zeng2021transporter, lin2020invariant}. Among them, RAD~\citep{laskin2020RAD} is considered the first work to systematically demonstrate that simple augmentation methods can significantly improve the sample efficiency and generalization. Other representative works, such as CURL~\citep{laskin2020curl}, introduce contrastive learning to align representations of differently augmented observations, while DrQ~\citep{yarats2021DrQ, yarats2021DrQ2} enforces consistency by averaging policy and value updates over multiple augmented views, regularizing the training objective and reducing overfitting and variance in target estimation.
In contrast to image-based RL, state-based RL has received relatively less attention.
Several approaches focus on generating or transforming state transitions while preserving dynamics consistency, including counterfactual and model-based augmentation methods~\citep{pitis2020counterfactual, pitis2022mocoda}, as well as guided or dynamics-invariant transformations for offline and online settings~\citep{corrado2023guided, corrado2024understanding}. EuclideanDA~\citep{luo2024euclidean} instead adopts a limb-based kinematic state representation to generate rich augmented data under Euclidean transformations. Notably, RAD~\citep{laskin2020RAD} also explores augmenting observations by additive Gaussian perturbations or random amplitude scaling. Although its improvements are generally minor and unstable in practice, it provides a simple and practical baseline for state-based data augmentation.

\section{Preliminaries}
\subsection{Problem Formulation}
We formulate reinforcement learning tasks as a Markov Decision Process (MDP), defined by the tuple 
$\mathcal{M} = (\mathcal{S}, \mathcal{A}, \mathcal{P}, r, \gamma)$, where $\mathcal{S}$ and $\mathcal{A}$ denote the state and action spaces,  $\mathcal{P}$ is the transition probability, $r$ is the reward function, and $\gamma$ is the discount factor. Throughout this paper, both $\mathcal{S}$ and $\mathcal{A}$ are assumed to be continuous. 
At each timestep $t$, the agent observes $s_t \in \mathcal{S}$, selects an action $a_t \in \mathcal{A}$ according to a policy $\pi_\theta(a_t|s_t)$, and receives a reward $r(s_t, a_t)$. The objective is to learn a policy that maximizes the expected discounted return $\mathbb{E}_{\pi}[\sum_{t=0}^\infty \gamma^t r(s_t, a_t)]$.

\subsection{Group-Invariant Markov Decision Processes}
In particular, a reflection operation induces a finite group of order two, commonly denoted as $\mathbb{Z}_2$, consisting of a non-trivial reflection element and the identity, with the group operation satisfying $g^2 = e$.
More generally, reflection symmetry can be viewed as a special case of group-invariant Markov decision processes (MDPs)~\citep{wang2022SO2}, in which the reward function and transition dynamics are preserved under the operation of a symmetry group $G$.

\begin{definition}[$G$-invariant MDP]
A $G$-invariant Markov decision process is defined as
$\mathcal{M}_G = (\mathcal{S}, \mathcal{A}, \mathcal{P}, r, \gamma, G)$.
For any $g \in G$, the induced group action on a state--action pair is given by:
\[
g \cdot (s, a) = (g s, g a).
\]
And it satisfies the following reward and transition invariant conditions:
\begin{align}
r(s, a) &= r(g s, g a), 
\label{eq:reward_invariance} \\
\mathcal{P}(s'\mid s, a)
&= \mathcal{P}(g s'\mid g s, g a).
\label{eq:transition_invariance}
\end{align}
\end{definition}

In this work, the group $G$ consists solely of non-identity reflection transformations on the state and action spaces.

\section{Symmetry in Continuous Control}
\label{sec:symmetry}
This section formalizes reflection symmetry in state-based continuous control and establishes its implications for optimality. We show that in a $G$-invariant MDP, the Bellman operator, optimal value functions and optimal policies all respect the underlying symmetry. These results provide a theoretical foundation for incorporating reflection symmetry into policy learning, which we exploit in Section~\ref{sec:method}.

\subsection{Reflection Symmetry in Continuous Control}
To model reflection symmetry in continuous control, let $\ell_u$ denote the reflection axis, and let $u \in \mathbb{R}^d$ be the unit normal vector perpendicular to $\ell_u$. The reflection matrix is defined as $R = I - 2uu^\top$, where $I$ is the identity matrix. Since various physical quantities transform differently under reflection, we decompose each state $s_t$ and action $a_t$ into semantically distinct components to ensure consistent and physically meaningful transformations.

\begin{figure*}[t]
    \centering

    \begin{subfigure}{0.45\linewidth}
        \centering
        \includegraphics[width=\linewidth]{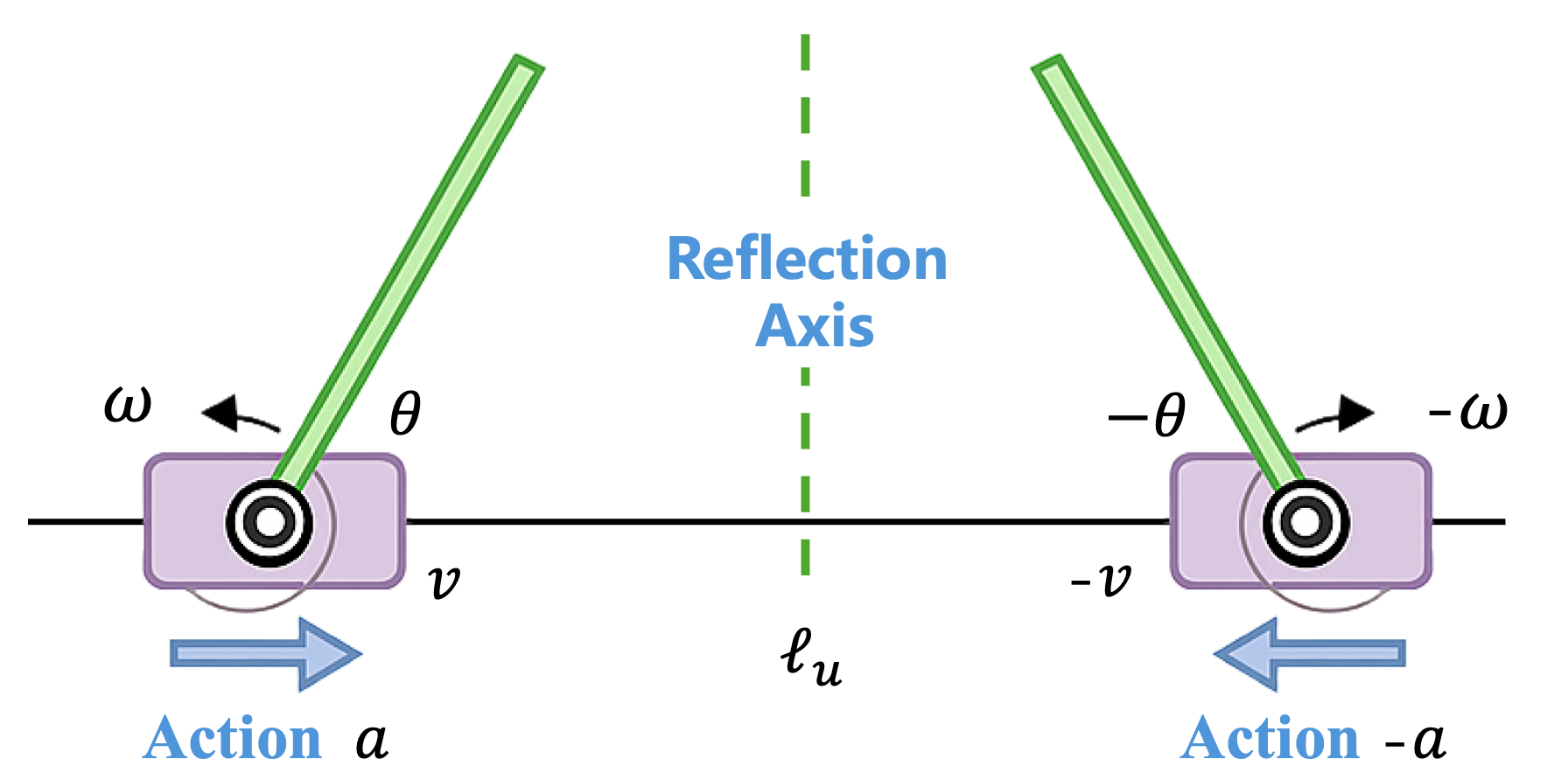}
        \caption{Axial Reflection}
        \label{fig:axi_ref}
    \end{subfigure}
    \hfill
    \begin{subfigure}{0.45\linewidth}
        \centering
        \includegraphics[width=\linewidth]{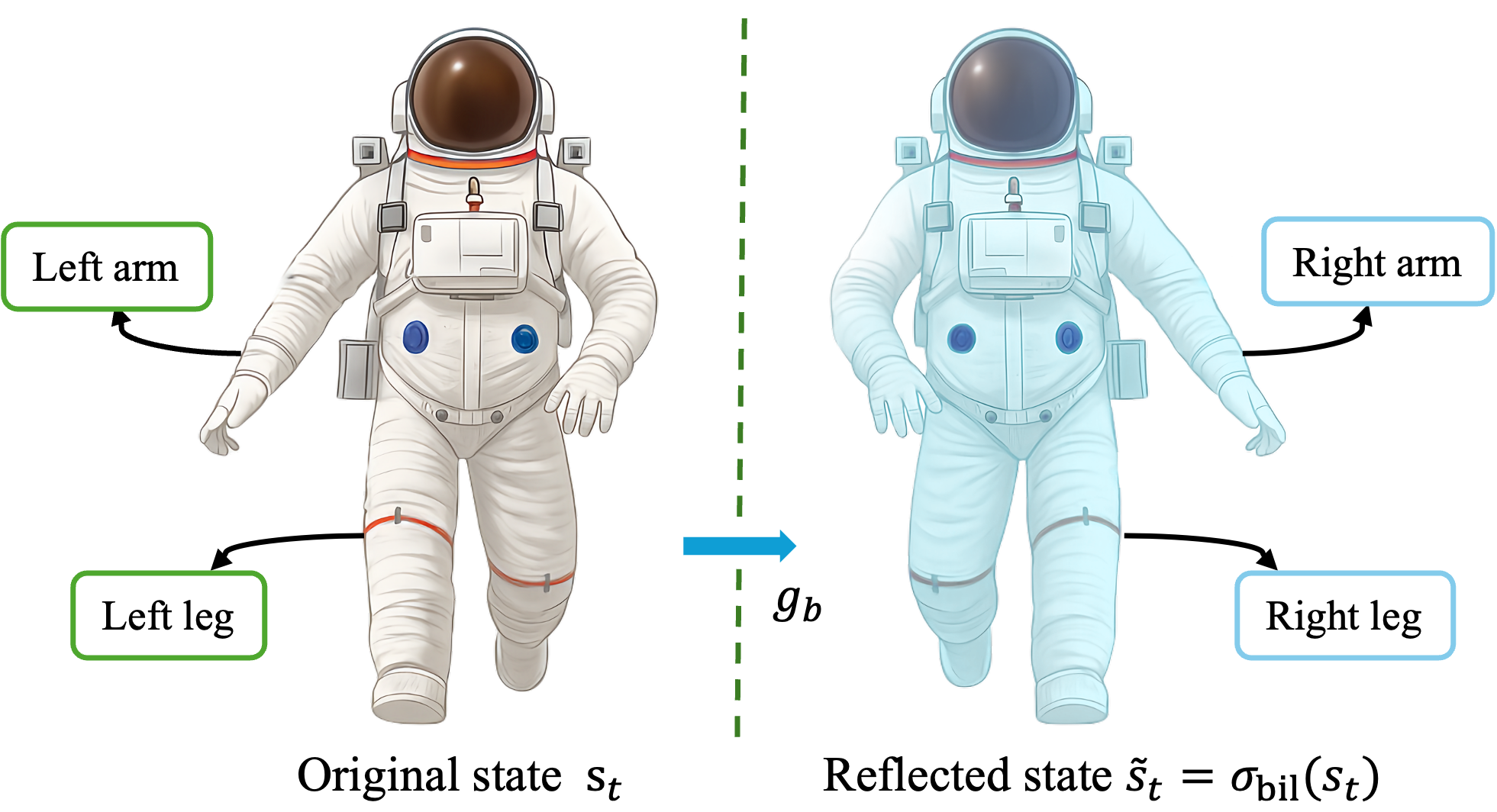}
        \caption{Bilateral Reflection}
        \label{fig:bil_ref}
    \end{subfigure}

    \caption{\textbf{Left:} A simple illustration of axial reflection. The cart moves along the $x$-axis, with state $(\theta,\omega,v)$ (pole angle, angular velocity, and cart velocity) and action $a$ (force to the cart). The reflection matrix to $\ell_u$ is $R= \begin{bsmallmatrix} -1 & 0 \\ 0 & 1 \end{bsmallmatrix}$. After axial reflection, the augmented state becomes $\tilde{s}=\bigl(-\theta, -\omega, -v\bigr)$ and the action is $-a$, preserving dynamics symmetry. \textbf{Right:} An illustration of bilateral reflection. Instead of a physical reflection over operation $g_b$, the astronaut swaps the gait information of the left and right sides, $\tilde{s}_{\text{left}} \leftrightarrow \tilde{s}_{\text{right}}$.}
\end{figure*}

% \begin{figure*}[t]
%   \centering
%   \subfigure[Axial Reflection]{\includegraphics[width=0.45\linewidth]{figures/axial_reflection.png}\label{fig:axi_ref}}
%   \subfigure[Bilateral Reflection]{\includegraphics[width=0.45\linewidth]{figures/bilateral_ref.png}\label{fig:bil_ref}}
%   \caption{\textbf{Left:} A simple illustration of axial reflection. The cart moves along the $x$-axis, with state $(\theta,\omega,v)$ (pole angle, angular velocity, and cart velocity) and action $a$ (force to the cart). The reflection matrix to $\ell_u$ is $R= [\begin{smallmatrix} -1 & 0 \\ 0 & 1 \end{smallmatrix}]$. After axial reflection, the augmented state becomes $\tilde{s}=\bigl(-\theta, -\omega, -v\bigr)$ and the action is $-a$, preserving dynamics symmetry. \textbf{Right:} An illustration of bilateral reflection. Instead of a physical reflection over operation $\sigma_\mathrm{bil}$, the astronaut swaps the gait information of the left and right sides $\tilde{s}_{\text{left}} \leftrightarrow \tilde{s}_{\text{right}}$.}
% \end{figure*}

% \begin{definition}[Axial Reflection]
% \label{def_axial}
% An axial reflection is a group element $g \in \mathrm{O}(2)$ acting on the $(s_t,a_t)$, which satisfies the conditions in Table~\ref{tab:reflection_rules}.
% \end{definition}

\begin{definition}[Axial Reflection]
An \textit{axial reflection} is a group element $g_a$ that acts on the state–action pair $(s_t,a_t)$, and $g_a$ induces a bijective transformation $g_a: (\mathcal{S}, \mathcal{A}) \to (\mathcal{S}, \mathcal{A})$ that follows the component-wise mappings in Table~\ref{tab:reflection_rules} and satisfies $g_a^2 = e$.
\end{definition}

\begin{table}[t]
\centering
\caption{
Reflection transformations applied to common state and action components. Variables with a tilde denote the transformed components under the reflection operation
}
\label{tab:reflection_rules}
% \begin{tabularx}{\columnwidth}{lclc}
% \toprule
% \textbf{Component} & \textbf{Example} & \textbf{Symbol} & \textbf{Transformation} \\
% \midrule
% Orientation & rotation matrix & ${O}$ & $\tilde{{O}} = R {O} R$ \\
% Euclidean & position, velocity, force & ${x}_e$ & $\tilde{{x}}_e = R {x_e}$ \\
% Pseudovector & angular velocity, torque & ${x}_p$ & $\tilde{{x}}_p = -R {x_p}$ \\
% Angular & orientation angle & $x_\theta$ & $\tilde{\theta} = \mathrm{atan2}(R x_\theta)$ \\
% Invariant & time, temperature & $x_q$ & $\tilde{x}_q = x_q$ \\
% \bottomrule
% \end{tabularx}
% \end{table}

\begin{tabularx}{\columnwidth}{lcc}
\toprule
\textbf{Component} & \textbf{Transformation} & \textbf{Example}\\
\midrule
Orientation ${O}$ & $\tilde{{O}} = R {O} R$ & rotation matrix \\
Euclidean ${x}_e$ & $\tilde{{x}}_e = R {x_e}$ & velocity, force \\
Pseudovector ${x}_p$ & $\tilde{{x}}_p = -R {x_p}$ & angular velocity \\
Angular ${x}_\theta$ & $\tilde{{x}}_\theta = \mathrm{atan2}(R {x}_\theta)$ & orientation angle \\
Invariant ${x}_q$ & $\tilde{{x}}_q = {x}_q$ & time, temperature \\
\bottomrule
\end{tabularx}
\end{table}

The derivation and detailed description of these transformations are provided in Appendix~\ref{appendix:trans}.
As illustrated in Figure~\ref{fig:axi_ref}, the CartPole environment provides a canonical example of axial reflection, corresponding to a reflection of the system about its vertical axis.

\begin{definition}[Bilateral Reflection] A bilateral reflection is a group element $g_b$ whose axis $\ell_u$ coincides with agent's intrinsic symmetry axis. And it satisfies:

{
  \setlength{\abovedisplayskip}{-4pt}
\setlength{\abovedisplayshortskip}{-4pt}
\begin{equation}
\label{eq:bilateral_reflection}
g_b \cdot (s, a) = \sigma(g_a s,\, g_a a)
\end{equation}
}

where $\sigma$ swaps the left and right symmetric components:
$\sigma(\tilde{s}_{\mathrm{left}},\tilde{s}_{\mathrm{right}},\tilde{a}_{\mathrm{left}},\tilde{a}_{\mathrm{right}})
\mapsto
(\tilde{s}_{\mathrm{right}},\tilde{s}_{\mathrm{left}},\tilde{a}_{\mathrm{right}},\tilde{a}_{\mathrm{left}})$.
\end{definition}

Intuitively, bilateral reflection can be  regarded as a special case of axial reflection. However, unlike general axial reflections, it does not correspond to a physically realizable spatial transformation with respect to a geometric axis, just as it is impossible to physically reflect a human body across its left–right axis. Instead, as illustrated in Figure~\ref{fig:bil_ref} (e.g., swapping the astronaut’s left and right limbs), it entails an abstract exchange of symmetric components within the state–action space. Clearly, bilateral reflection is only applicable to agents with an intrinsically mirrored architecture. 

\subsection{Symmetry of Optimal Policy}
\begin{lemma}[Equivariance of the Bellman Operator]
\label{lemma:bellman}
Let $\mathcal{T}$ denote the Bellman optimality operator. For any value function $V$ and any $g \in G$ in G-invariant MDP $\mathcal{M}_G$:
\[
\mathcal{T}(V \circ g) = (\mathcal{T}V) \circ g.
\]
\end{lemma}

The proof is provided in Appendix~\ref{app:proof_bellman}.
Lemma~\ref{lemma:bellman} establishes that the Bellman optimality operator commutes with the group action induced by a $G$-invariant MDP.
In particular, the equivariance of the Bellman operator implies that symmetry present in a value function is preserved under Bellman updates.
As a consequence, the unique fixed point of the Bellman optimality operator, i.e., the optimal value function $V^*$, must be $G$-invariant.

\begin{lemma}[Invariance of Optimal Value Functions]
\label{lemma:value_invariance}
In a $G$-invariant MDP $\mathcal{M}_G$, the optimal state-value function $V^*$ and action-value function $Q^*$ are invariant under the group action $G$. For any $g \in G$, $s \in \mathcal{S}$, and $a \in \mathcal{A}$:
\[
V^*(s) = V^*(g s), \quad Q^*(s, a) = Q^*(g s, g a),
\]
\end{lemma}

The proof is provided in Appendix~\ref{app:value_invariance}.
Lemma~\ref{lemma:value_invariance} shows that the optimal value functions $V^*$ and $Q^*$ are invariant under the group action in a $G$-invariant MDP.
In particular, the invariance $Q^*(s,a)=Q^*(g s, g a)$ implies that optimal actions are mapped consistently across group-related states.
This structural property allows symmetry at the value-function level to be lifted to the policy level, leading to the following theorem on the equivariance of optimal policies.

\begin{theorem}[Equivariance of Optimal Policies]
\label{theorem:policy_equi}
In a $G$-invariant MDP $\mathcal{M}_G$, there exists an optimal policy $\pi^*$ that is $G$-equivariant.
Specifically, for a stochastic policy, equivariance is characterized by:
\[
\pi^*(a \mid g s) = \pi^*(g^{-1} a \mid s),
\quad \forall g \in G,\; s \in \mathcal{S},\; a \in \mathcal{A}.
\]
For a deterministic policy $a=\mu^*(s)$, equivariance condition is given by:
\[
\mu^*(g s) = g \mu^*(s), \quad \forall g \in G,\; s \in \mathcal{S}.
\]
\end{theorem}

The proof is provided in Appendix~\ref{app:policy_equi}.
Theorem~\ref{theorem:policy_equi} shows that in a $G$-invariant MDP, symmetry is preserved not only by the optimal value functions but also by the optimal policies themselves.
In the continuous-action setting considered in this work, policies are typically parameterized as Gaussian distributions $\pi^(a \mid s) = \mathcal{N}\big(m^(s), \Sigma^(s)\big)$.
By the invariance of the optimal value functions established in Lemma~\ref{lemma:value_invariance}, the policy parameters inherit the same equivariance structure, yielding
$ m^*(gs) = gm^*(s)$ and $\Sigma^*(gs) = g\,\Sigma^*(s)\,g^\top$.
This result confirms that the resulting Gaussian policy is $G$-equivariant as well.

This result provides a rigorous theoretical foundation for exploiting symmetry in policy learning, ensuring that enforcing equivariance does not compromise optimality. Motivated by this property, we next develop practical and scalable mechanisms that explicitly incorporate reflection symmetry into policy learning, with the goal of improving both final performance and sample efficiency.

\section{Methodology}
\label{sec:method}
In this section, we present the integration of reflection symmetry into specific RL algorithms. In particular, we select PPO and SAC as representatives. Implementation details of the resulting algorithms are provided in Appendix~\ref{app:algorithms}.

\subsection{Reflex PPO}
For PPO, we follow prior approaches~\cite{yu2018learning,abdolhosseini2019Onlearning,mittal2024symmetry}, which typically incorporate explicit auxiliary regularization terms to the learning objective.
Motivated by the theoretical results in Section~\ref{sec:symmetry}, we introduce a reflection symmetry regularization that explicitly enforces equivariance in the learned policy:

\begin{equation}
    \label{eq:reflex-ppo_Aloss}
    \mathcal{L}^{\text{sym}}_{\pi}(\theta) = 
    \mathbb{E}_{s_t \sim \tau,\, g \sim {G}} 
    \bigl\|\,\pi_\theta(g s_t) - g(\pi_\theta(s_t))\,\bigr\|_2^2.
\end{equation}

Intuitively, this regularization promotes information sharing between symmetric components, allowing the policy to leverage mirrored experiences without additional environment interaction. 
However, most existing PPO-based symmetry-aware approaches focus exclusively on enforcing consistency at the policy level, leaving the value function unconstrained. This can result in asymmetric value estimates across symmetric states, which in turn propagate unnecessary variance and instability into policy updates, we empirically verify in Section~\ref{sec:ablation}.
To address this issue, we additionally regularize the critic by enforcing value consistency following~\ref{lemma:value_invariance}:
\begin{equation}
    \label{eq:reflex-ppo_Closs}
    \mathcal{L}^{\text{sym}}_{V}(\phi) = 
    \mathbb{E}_{s_t \sim \tau,\, g \sim {G}} 
    (V_\phi(g s_t) - V_\phi(s_t))^2.
\end{equation}

And combined with the standard PPO actor and critic loss:
\begin{align}
\mathcal{L}_V(\phi)
&=\mathbb{E}_t\!\left[\bigl(V_\phi(s_t)-V^{\text{tgt}}_t\bigr)^2\right],
\label{eq:ppo_Closs}\\
\mathcal{L}_{\pi}(\theta)
&=\mathbb{E}_t\!\left[
\min\!\Big(
\rho_t(\theta)\hat{A}_t,\,
\mathrm{clip}\!\big(\rho_t(\theta),1-\epsilon,1+\epsilon\big)\hat{A}_t
\Big)
\right].
\label{eq:ppo_Aloss}
\end{align}

where $\rho_t(\theta) = \frac{\pi_\theta(a_t \mid s_t)}{\pi_{\theta'}(a_t \mid s_t)}$ denotes the importance sampling ratio,
$\epsilon$ is the clipping parameter, and $\hat{A}_t$ is the advantage estimate computed using Generalized Advantage Estimation (GAE)~\cite{schulman2015high}.
The overall training objective thus becomes a weighted combination of the original PPO loss 
and the symmetry-consistency regularizers:
\begin{equation}
    \mathcal{L}_{\text{total}} = 
    \mathcal{L}_{\pi}  + \mathcal{L}_V
    + w_t(
    \mathcal{L}^{\text{sym}}_{\pi} + 
    \mathcal{L}^{\text{sym}}_{V}),
\end{equation}
where $w_t$ controls the strength of the symmetry regularization.
We innovatively linearly decay it over training steps as 
$w_t = w_0 (1 - t / T)$, 
where $t$ denotes the current timestep and $T$ is the total timesteps.
This schedule imposes a strong inductive bias early in training to promote information sharing across symmetric states and stabilize optimization, while gradually relaxing the constraint to allow more flexible policy adaptation. As shown in Section~\ref{sec:ablation}, the proposed decaying schedule outperforms a fixed regularization weight.

\subsection{Reflex SAC}
For SAC, we adopt a \emph{symmetry regularization} strategy inspired by the philosophy of DrQ~\citep{yarats2021DrQ}.
While DrQ improves sample efficiency by applying stochastic augmentations to image observations,
our approach instead exploits reflection symmetry inherent in structured state vector representations.
Given a minibatch $\mathcal{B} = \{(s_i, a_i, r_i, s'_i)\}_{i=1}^{|\mathcal{B}|}$ sampled from the replay buffer $\mathcal{R}$,
the standard target for critic learning is defined as:
\begin{equation}
\label{eq:sac_target}
y = r(s,a) + \gamma V(s'),
\end{equation}
where the soft state value $V(s')$ is given by:
\begin{equation}
V(s') =
\mathbb{E}_{a' \sim \pi_\theta(\cdot \mid s')}
\Big[
\min_{i=1,2} Q_{\bar{\phi}_i}(s',a')
- \alpha \log \pi_\theta(a' \mid s')
\Big].
\end{equation}

To incorporate reflection symmetry, we additionally construct a reflected target.
Specifically, we apply a group transformation $g \in G$ to obtain the mirrored next state
$\tilde{s}' = g s'$,
and sample a corresponding action
$\tilde{a}' \sim \pi_\theta(\cdot \mid \tilde{s}')$.
The reflected target $\tilde{y}$ is then computed by evaluating Eq.~\eqref{eq:sac_target}
at $(\tilde{s}', \tilde{a}')$.
The final target used for critic regression is defined as the average of the original and reflected targets:
\begin{equation}
\bar{y} = \tfrac{1}{2}\bigl(y + \tilde{y}\bigr).
\label{eq:sac_ref_target}
\end{equation}

This symmetric target enforces consistency of value estimates across mirrored state--action pairs, thereby encouraging reflection invariance in the learned $Q$-function. Finally, we define the symmetry-regularized critic loss as:
\begin{equation}
\mathcal{L}_Q
= \mathbb{E}_{(s,a)\sim \mathcal{B}}
\!\left[
\bigl(Q_\phi(s,a) - \bar{y}\bigr)^2
+
\bigl(Q_\phi(gs,ga) - \bar{y}\bigr)^2
\right].
\label{eq:sac_sym_q}
\end{equation}

Because the SAC actor is optimized with respect to the learned $Q$-function, symmetry enforced at the critic level naturally induces equivariant behavior in the policy, consistent with our theoretical analysis. Hence, we do not further regularize the actor in SAC.

\section{Experiments}
\label{sec:Experiment}

\begin{table*}[t]
    \centering
    \small % 稍微缩小字体以适应较多列
    \caption{Performance and sample efficiency comparison across tasks.
\%$\uparrow$ denotes the percentage improvement of Reflex over the baseline. Sample efficiency is measured by the \emph{relative} number of environment steps required to reach the baseline’s final performance. Symbol "\texttt{-}" denotes the final performance is reached limit performance of the environment.}
    \label{tab:result}
    \setlength{\tabcolsep}{4pt} % 调整列间距，防止溢出
    \begin{tabular}{l ccc ccc | cc cc}
        \toprule
        & \multicolumn{6}{c}{\textbf{Final Performance}} & \multicolumn{4}{c}{\textbf{Steps to Baseline Final}} \\
        \cmidrule(lr){2-7} \cmidrule(lr){8-11}
        & \multicolumn{3}{c}{PPO-based} & \multicolumn{3}{c}{SAC-based} & \multicolumn{2}{c}{PPO-based} & \multicolumn{2}{c}{SAC-based} \\
        \cmidrule(lr){2-4} \cmidrule(lr){5-7} \cmidrule(lr){8-9} \cmidrule(lr){10-11}
        Environment & PPO & Reflex & \%$\uparrow$ & SAC & Reflex & \%$\uparrow$ & PPO & Reflex & SAC & Reflex \\
        \midrule
        CartpoleBalance  & 795.7$\pm$153.3  & 893.2$\pm$6.1 & 12\% & 995.6$\pm$0.8 & 996.7$\pm$0.5 & -- & 0.67 & 0.49 & 0.65 & 0.37 \\
        CartThreePoles   & 144.4$\pm$1.3 & 154.8$\pm$6.9 & 7\% & 174.8$\pm$4.6 & 190.2$\pm$8.8 & 9\% & 0.72 & 0.40 & 0.64 & 0.26 \\
        IDPendulum       & 5893.6$\pm$259.2 & 6174.6$\pm$770.7 & 5\% & 9343.1$\pm$1.1 & 9347.1$\pm$0.9 & -- & 0.79 & 0.54 & 0.40 & 0.36 \\
        WalkerRun        & 147.6$\pm$56.5 & 182.5$\pm$57.4 & 24\% & 753.6$\pm$19.6 & 809.6$\pm$6.2 & 7\% & 1.00 & 0.63 & 1.00 & 0.11 \\
        Walker2d         & 2724.1$\pm$889.8 & 3672.5$\pm$338.7 & 35\% & 4488.0$\pm$503.6 & 5042.9$\pm$242.5 & 12\% & 1.00 & 0.41 & 1.00 & 0.44 \\
        Ant-v3           & 2808.7$\pm$1016.8 & 3729.2$\pm$665.7 & 33\% & 4755.8$\pm$1892.2 & 5375.0$\pm$541.4 & 13\% & 1.00 & 0.48 & 1.00 & 0.40 \\
        \bottomrule
    \end{tabular}
\end{table*}

\begin{figure*}[t]
    \centering

    \begin{subfigure}{0.33\textwidth}
        \centering
        \includegraphics[width=\linewidth]{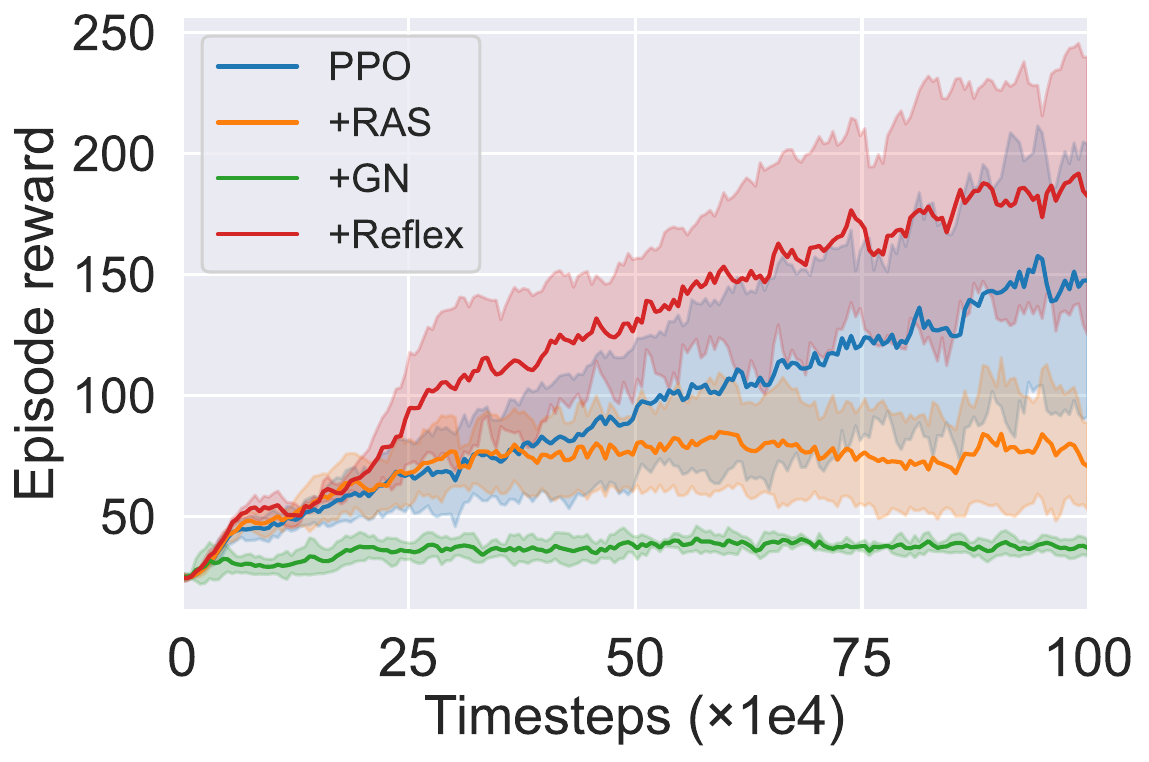}
        \caption{PPO-WalkerRun}
    \end{subfigure}
    \hfill
    \begin{subfigure}{0.33\textwidth}
        \centering
        \includegraphics[width=\linewidth]{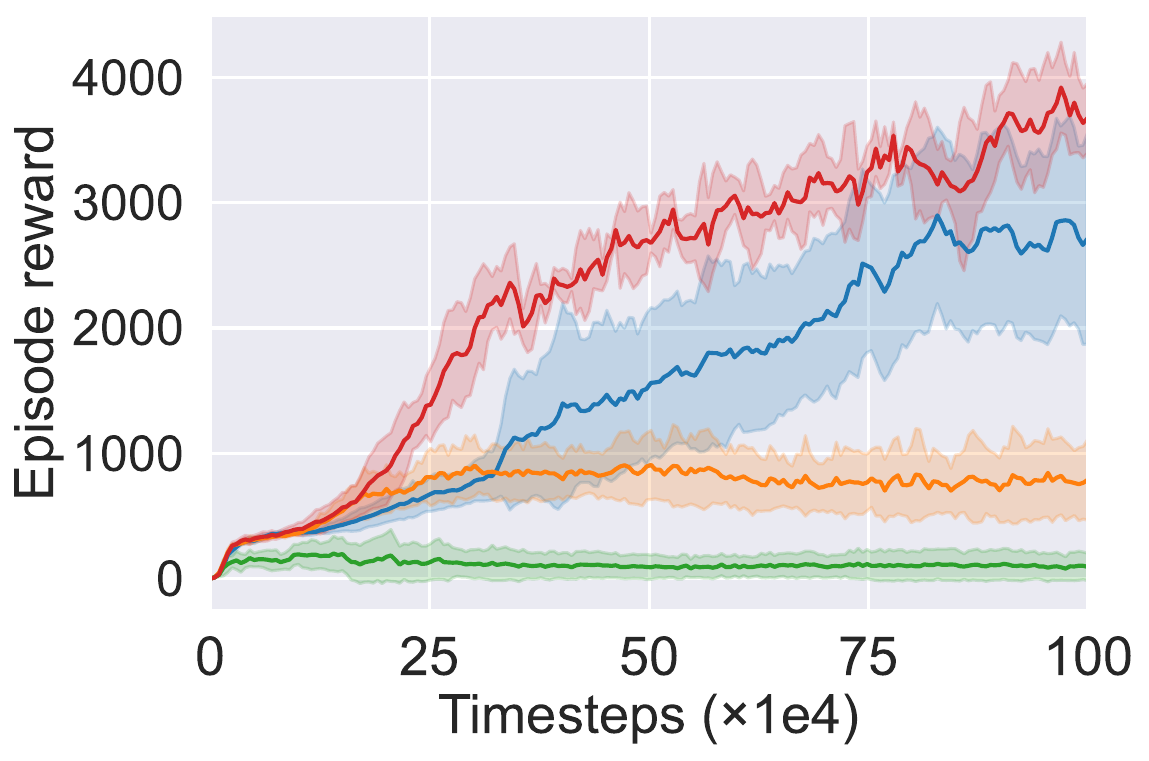}
        \caption{PPO-Walker2d}
    \end{subfigure}
    \hfill
    \begin{subfigure}{0.33\textwidth}
        \centering
        \includegraphics[width=\linewidth]{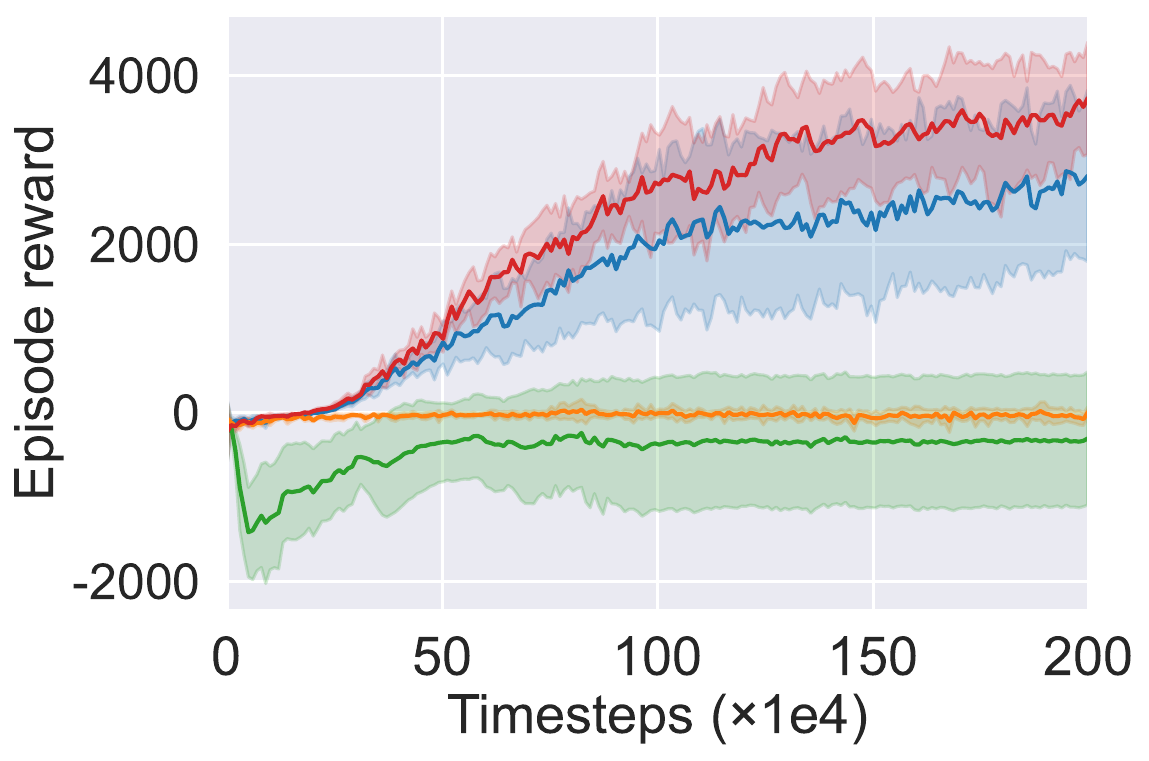}
        \caption{PPO-Ant}
    \end{subfigure}

    \begin{subfigure}{0.33\textwidth}
        \centering
        \includegraphics[width=\linewidth]{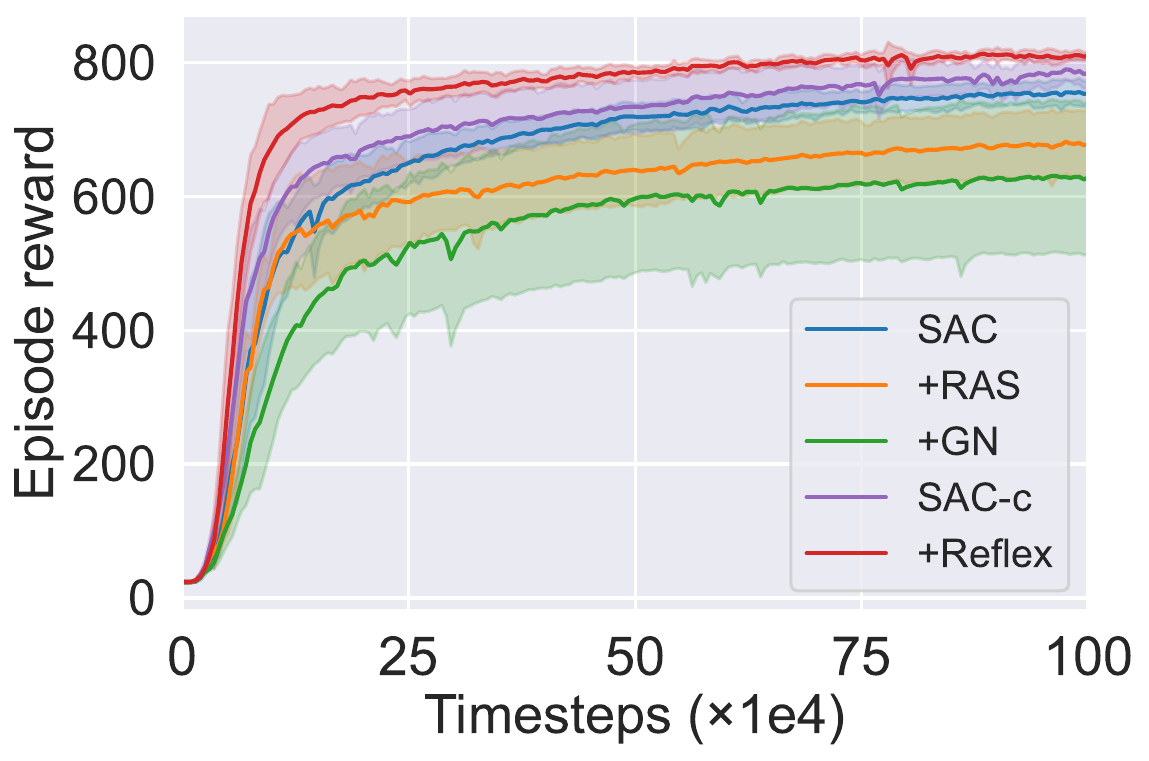}
        \caption{SAC-WalkerRun}
    \end{subfigure}
    \hfill
    \begin{subfigure}{0.33\textwidth}
        \centering
        \includegraphics[width=\linewidth]{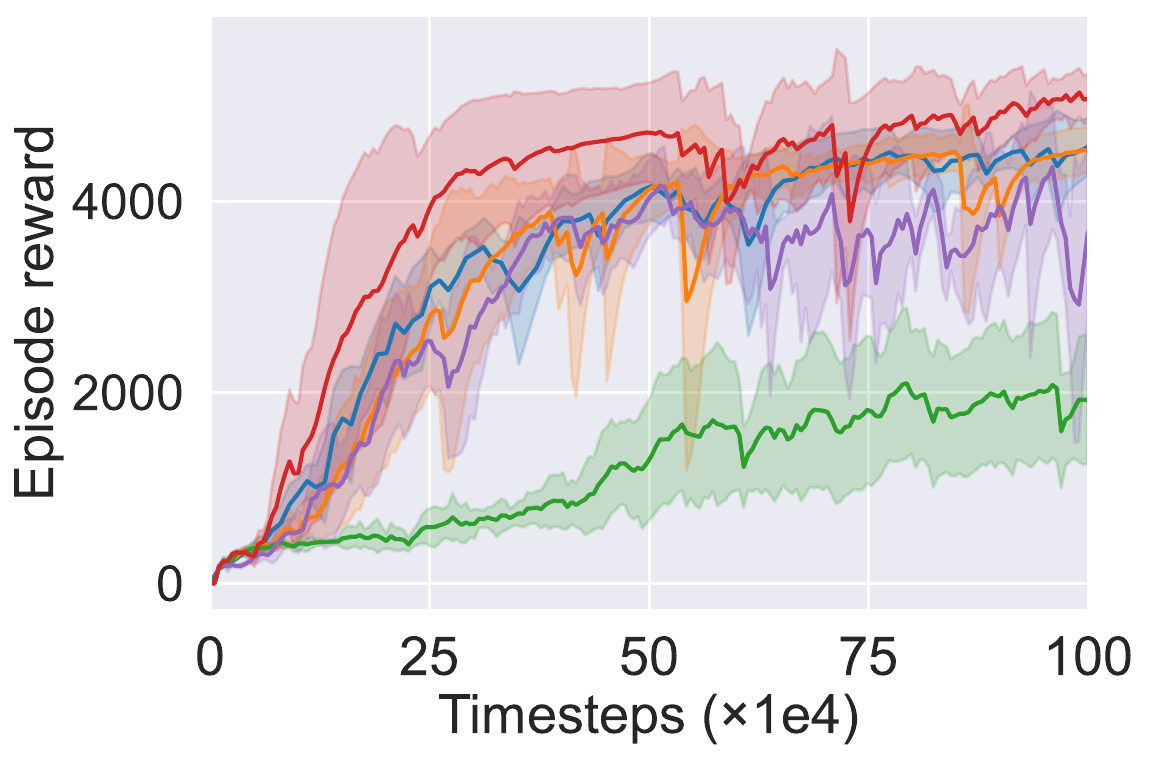}
        \caption{SAC-Walker2d}
    \end{subfigure}
    \hfill
    \begin{subfigure}{0.33\textwidth}
        \centering
        \includegraphics[width=\linewidth]{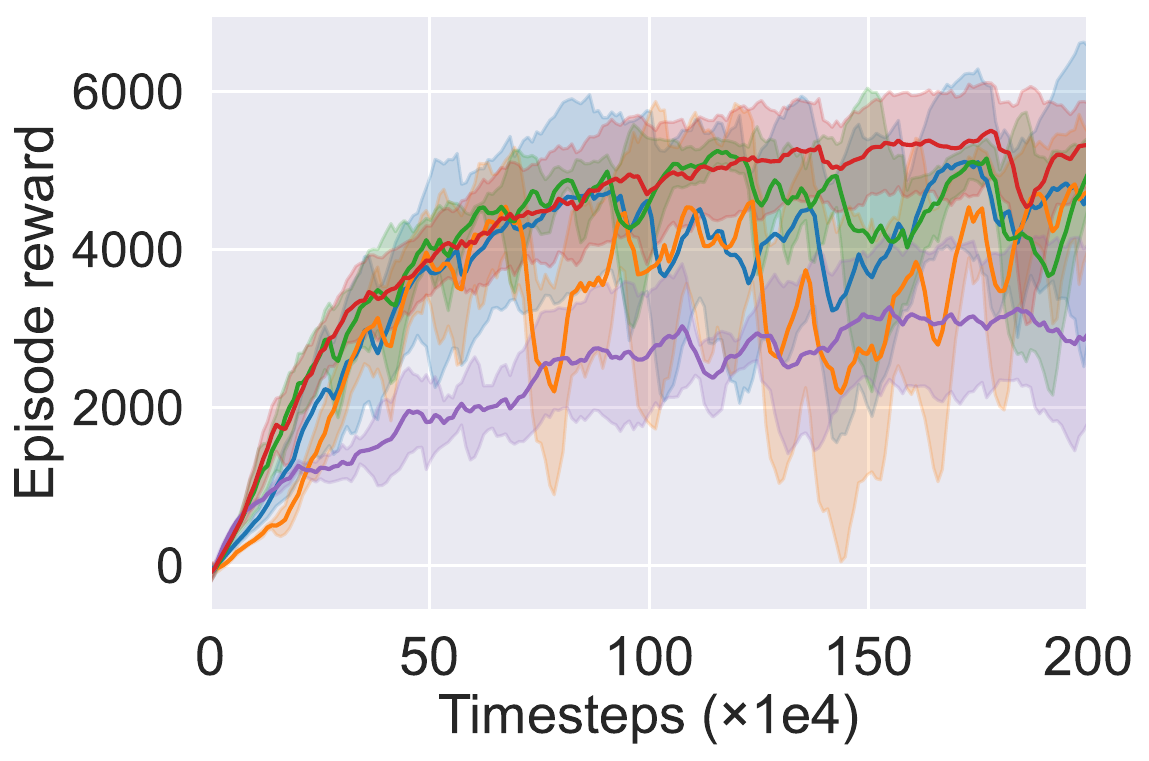}
        \caption{SAC-Ant}
    \end{subfigure}

    \caption{Learning curves of Reflex compared with baselines, where $w$ is set to 0.1 in Reflex-PPO. Due to space limitations, we report results only on tasks with bilateral reflection symmetry, complete results are provided in Appendix~\ref{app:full_results}.}
    \label{fig:result}
\end{figure*}

We conduct a comprehensive empirical evaluation of Reflex to assess its effectiveness, robustness, and generality across multiple RL settings. Specifically, our experiments are organized to answer the following questions:
\begin{itemize}[leftmargin=10pt, itemsep=0pt, topsep=2pt]
    \item Does Reflex consistently improve performance and sample efficiency?
% We evaluate Reflex integrated with PPO and SAC on a diverse suite of continuous control tasks, comparing against standard baselines and commonly used augmentation methods. 
(See Section~\ref{sec:main_result})
    \item Are all components necessary for Reflex performance gains?
% Through systematic ablation studies on Reflex-PPO, we analyze the individual contributions of actor, critic regularization, and the decaying symmetry weight to overall performance. 
(See Section~\ref{sec:ablation})
    \item How sensitive is Reflex to hyperparameter choices?
% We investigate the impact of key hyperparameters in both Reflex-PPO and Reflex-SAC, including symmetry regularization strength, critic target update frequency, learning rate and batch size, to characterize robustness across different optimization regimes.
(See Section~\ref{sec:hyp_analysis})
    \item How well does Reflex transfer to other RL algorithms beyond PPO and SAC?
% We examine the applicability of Reflex to TD3, demonstrating that reflection symmetry exploitation can be broadly beneficial beyond the primary methods considered.
(See Section~\ref{sec:applicability})
\end{itemize}

\subsection{Experimental Settings.}
\paragraph{Environments.} 
We evaluate our method on a diverse set of continuous control tasks from OpenAI Gymnasium~\citep{brockman2016gym} and the DeepMind Control Suite~\citep{tunyasuvunakool2020dm_control}, all implemented with the MuJoCo physics simulator~\citep{todorov2012mujoco}.
We select environments exhibiting either axial or bilateral reflection symmetry, as summarized in Table~\ref{tab:envs}. 
The specific description for each task are provided in the Appendix~\ref{app:env}.
For each task, we manually verify that the reward and transition dynamics satisfy the symmetry invariance.

\begin{table}[ht]
\centering
\caption{Summary of evaluation tasks.}
\label{tab:envs}
\renewcommand{\arraystretch}{1.15}
\setlength{\tabcolsep}{6pt}
\begin{tabular}{lcc}
\toprule
\textbf{Task} & \textbf{Source} & \textbf{Reflection Type} \\
\midrule
cart-threepoles-v0 & DMControl & Axial \\
cartpole-balance-v0 & DMControl & Axial \\
IDPendulum-v4 & Gymnasium & Axial \\
Walker2d-v4 & Gymnasium & Bilateral \\
Ant-v4 & Gymnasium & Bilateral \\
walker-run-v0 & DMControl & Bilateral \\
\bottomrule
\end{tabular}
\end{table}

\paragraph{Baselines.} We adopt PPO~\cite{schulman2017ppo} and SAC~\cite{haarnoja2018sac} as our on-policy and off-policy base algorithm respectively.
For comparison, we include RAD \cite{laskin2020RAD}, which introduces two general augmentation techniques for state-based RL: 
1) \textbf{Random Amplitude Scaling (RAS)}: randomly scales state vectors by a uniform random variable; and 
2) \textbf{Gaussian Noise (GN)}: adds small Gaussian noise to state vectors. 
We adopt the same parameters as reported in the original RAD implementation for all our experiments.
For RAS, the scaling factor $z$ is sampled from a uniform distribution $\text{Uni}[\alpha,\beta]$, 
where the minimum value $\alpha \in \{0.6, 0.8\}$ and the maximum value $\beta \in \{1.2, 1.4\}$. 
For GN, we perturb states as $s' = s + \mathbf{z}$, where $\mathbf{z} \sim \mathcal{N}(0, I)$. 
For Reflex-SAC, we set the critic target update frequency to 2 and adopt a more aggressive learning rate of $1\mathrm{e}{-3}$ with a batch size of 512. To ensure a fair and controlled comparison, we additionally evaluate a baseline SAC variant under the same hyperparameter configuration, denoted as \textbf{SAC-c} in Figure~\ref{fig:result}.

\paragraph{Training Details.} 
Our implementation is based on the CleanRL\footnote{
\url{https://github.com/vwxyzjn/cleanrl}
}, a     widely adopted open-source reinforcement learning repository.
We evaluate all methods every 2{,}000 environment steps by averaging episodic returns over 10 evaluation episodes with different random seeds. To ensure fair comparison, all algorithms are evaluated on the same set of 10 fixed evaluation environments at each evaluation point. Learning curves report the mean performance over 5 random seeds, with shaded regions indicating one standard deviation.
For Reflex-PPO, the symmetry regularization weight $w$ is set to $0.1$ by default.
% For Reflex-SAC, the  $w$ is set to $0.1$ by default.
Due to the lightweight neural architectures commonly used in state-based RL and the dominance of environment interaction costs, all experiments are conducted on a single Apple Mac M4 machine equipped with 10 CPU cores and 32GB RAM. This setup provides competitive training speed in practice. A complete list of hyperparameters is provided in Appendix~\ref{app:hyperparameters}.

\subsection{Results}
\label{sec:main_result}
Our experimental results are summarized in Table~\ref{tab:result} and Figure~\ref{fig:result}.
As shown in Figure~\ref{fig:result}, Reflex consistently outperforms all baselines across tasks. In contrast, generic augmentation strategies such as Gaussian Noise (GN) and Random Amplitude Scaling (RAS) exhibit unstable behavior and even degrade performance, particularly for PPO. We attribute this to the use of running mean and standard deviation normalization in state-based on-policy RL algorithms~\cite{raffin2021stable}, where perturbations introduced by GN and RAS can be further amplified, leading to distributional mismatch and degraded learning stability. Meanwhile, \texttt{SAC-c} exhibits unstable performance under same configuration, indicating that simply using a higher learning rate and less frequent target updates is ineffective. In contrast, reflection symmetry regularization stabilizes training and enables Reflex-SAC to achieve efficient and robust learning.

Table~\ref{tab:result} provides a quantitative comparison from two perspectives: \emph{final performance} and \emph{sample efficiency}. Final performance is measured by the accumulated episodic reward, while sample efficiency is quantified by the relative environment steps required for Reflex to reach the baseline’s final performance. As shown, Reflex achieves substantial gains in sample efficiency across all tasks, along with consistent improvements in final performance. Notably, the benefits are more pronounced in bilateral reflection tasks, where PPO-based methods achieve up to approximately 30\% improvement in final performance. These results demonstrate that explicitly exploiting reflection symmetry yields robust and scalable benefits in state-based continuous control.

\subsection{Ablation Study}
\label{sec:ablation}
We focuses on Reflex-PPO components and consider three variants: 1) \texttt{w/o A}, which removes the actor symmetry regularization;
2) \texttt{w/o C}, which removes the critic symmetry regularization; 
3) \texttt{w/o D}, which uses a fixed symmetry regularization weight $w$ instead of a decaying schedule.

\begin{figure}[ht]
    \centering
    \begin{subfigure}{0.49\columnwidth}
        \centering
        \includegraphics[width=\linewidth]{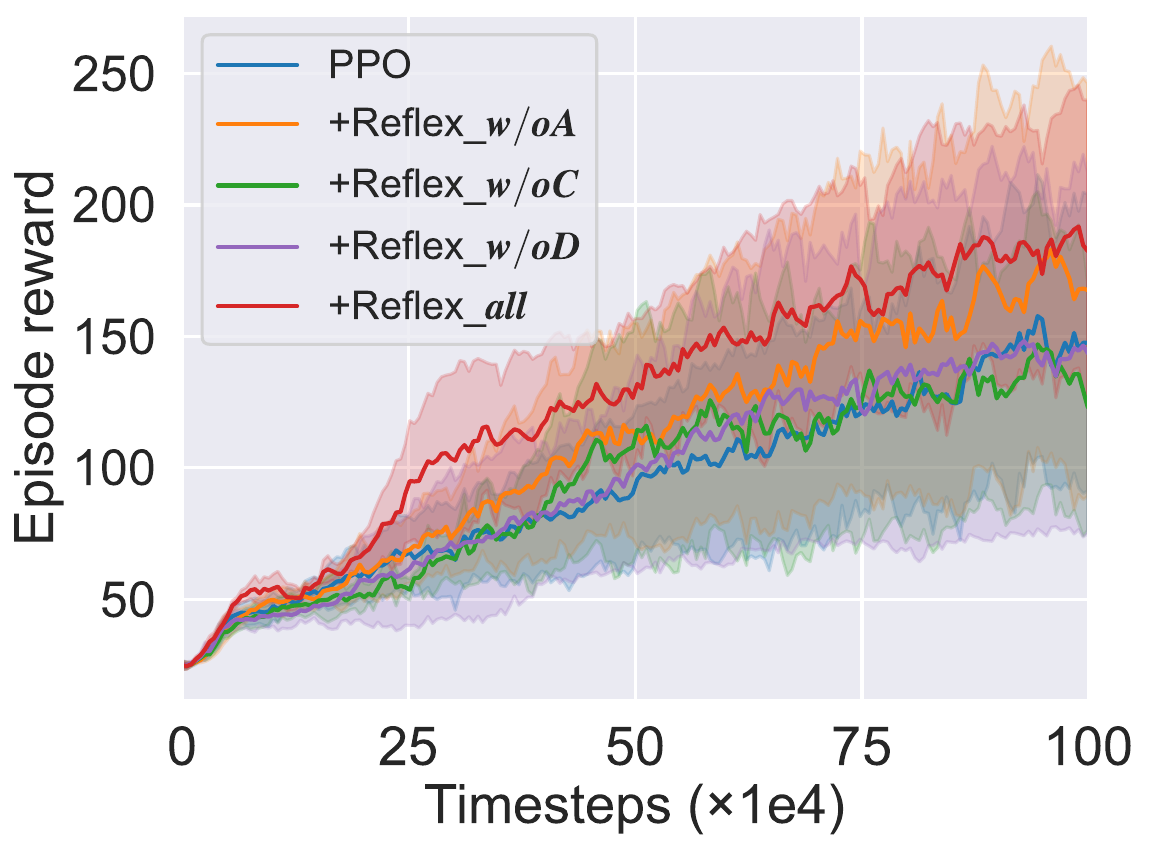}
        \caption{WalkerRun}
    \end{subfigure}
    \hfill
    \begin{subfigure}{0.49\columnwidth}
        \centering
        \includegraphics[width=\linewidth]{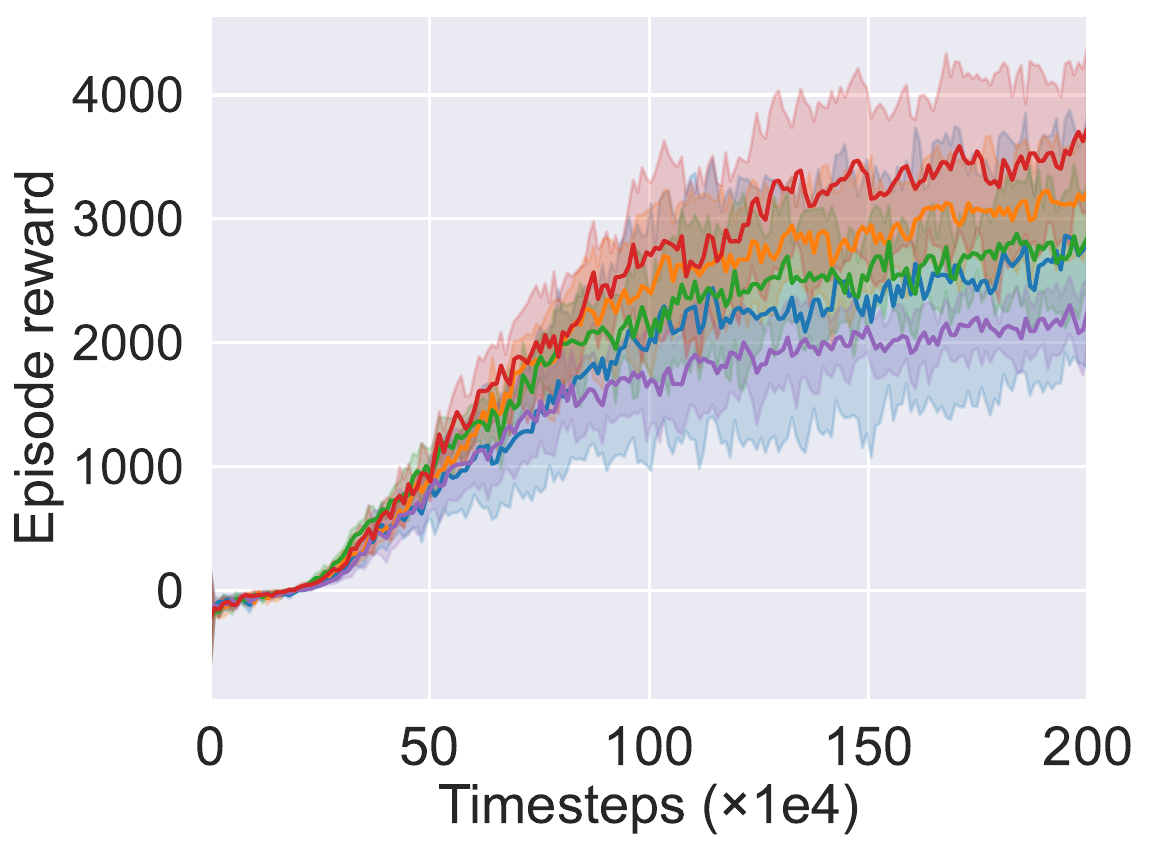}
        \caption{Ant}
    \end{subfigure}

    \caption{Ablation study result on Reflex-PPO, complete results are provided in Appendix~\ref{app:ablation}.}
    \label{fig:ablation}
\end{figure}

The results are presented in Figure~\ref{fig:ablation}. We observe that removing any symmetry regularization leads to significant performance degradation. Notably, omitting the critic regularization (\texttt{w/o C}) results in the most pronounced drop, underscoring the critical role of value consistency in stabilizing policy updates. Overall, these findings validate the design choices in Reflex-PPO and emphasize the synergistic benefits of actor and critic symmetry regularization combined with a dynamic weighting strategy.

\subsection{Comparison with Different Settings}
\label{sec:hyp_analysis}
\paragraph{Reflex-PPO.}
For Reflex-PPO, we further evaluate the effect of the symmetry regularization weight $w$ in the PPO experiments. 
The coefficient $w$ controls the relative strength of the symmetry loss with respect to the standard PPO objective. 
We tested multiple values of $w \in \{0.05, 0.1, 0.25, 0.5\}$ across bilateral reflection tasks to analyze its influence on learning performance.

\begin{figure}[ht]
    \centering
    \begin{subfigure}{0.49\columnwidth}
        \centering
        \includegraphics[width=\linewidth]{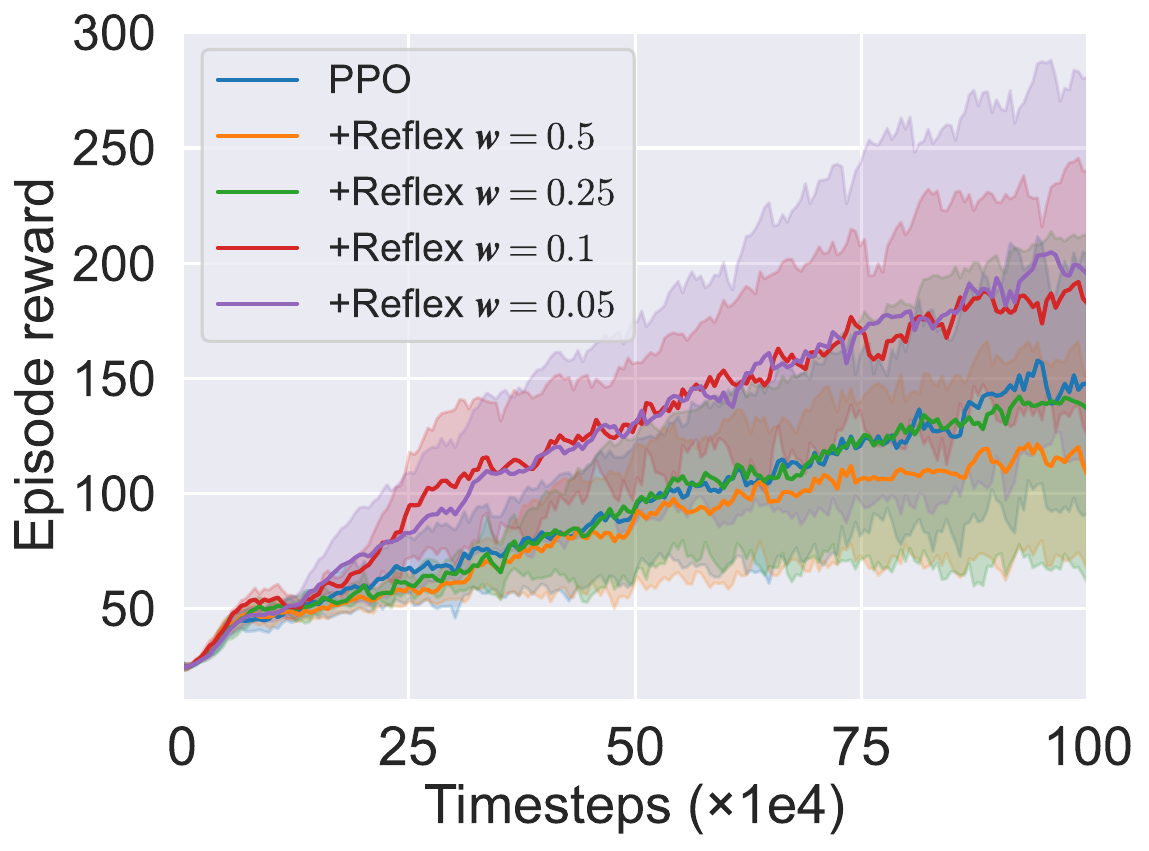}
        \caption{WalkerRun}
    \end{subfigure}
    \hfill
    \begin{subfigure}{0.49\columnwidth}
        \centering
        \includegraphics[width=\linewidth]{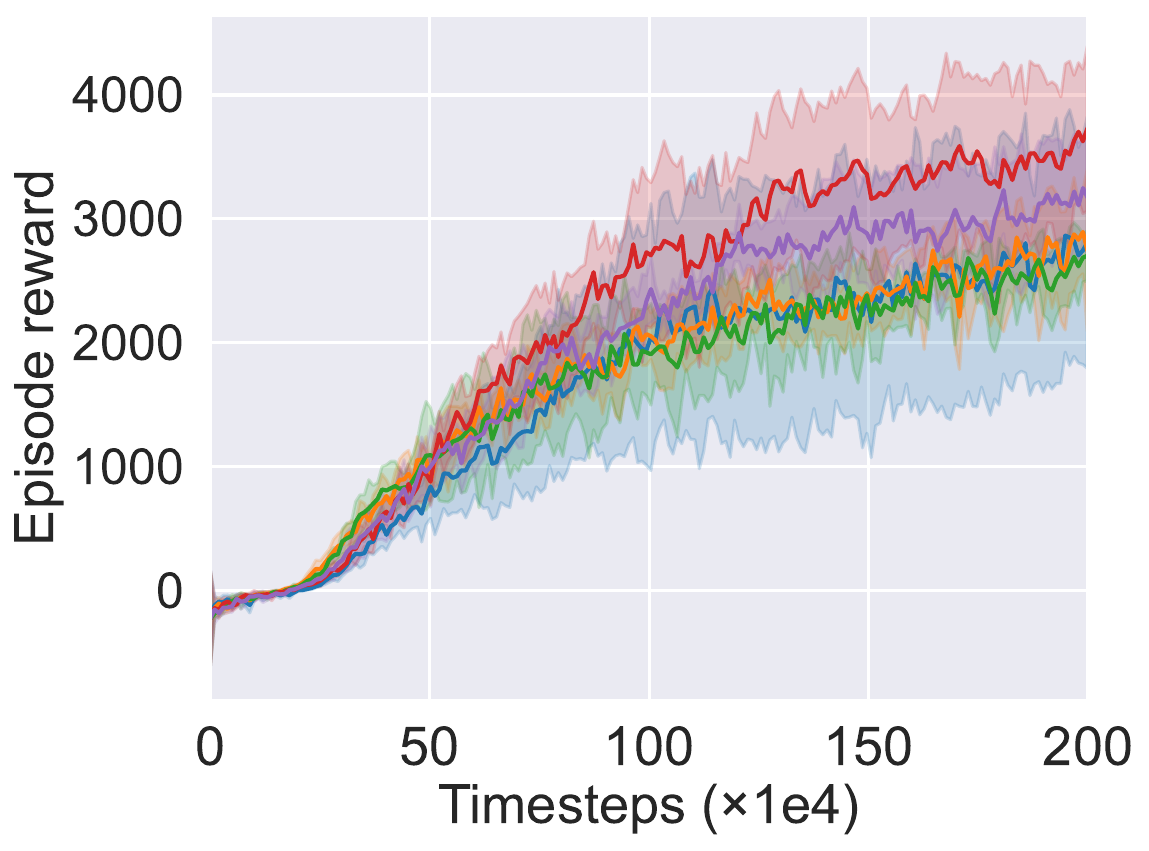}
        \caption{Ant}
    \end{subfigure}

    \caption{Sensitivity analysis of the regularization weight $w$ in Reflex-PPO, complete results are provided in Appendix~\ref{app:w_analysis}.}
    \label{fig:PPO_w}
\end{figure}

As shown in Figure~\ref{fig:PPO_w}, the performance is sensitive to the symmetry regularization weight $w$. 
Among all evaluations, $w=0.1$ consistently achieves the best results. 
A slightly smaller weight $w=0.05$ leads to a mild performance drop, suggesting that the regularization signal becomes insufficient. 
In contrast, larger weights rapidly degrade performance: when $w \ge 0.25$, the learning curves exhibit clear instability, 
and $w=0.25$ and $0.5$ remain the worst-performing settings throughout training. 
These results highlight that excessive symmetry constraints can dominate the PPO objective and hinder policy optimization, 
while a small but non-zero weight provides the most effective balance between enforcing symmetry and maintaining learning flexibility.

\paragraph{Reflex-SAC.}
For Reflex-SAC, we investigate the effect of different hyperparameters in the SAC experiments. Specifically, we evaluate the impact of varying the critic target update frequency $K \in \{1, 2, 3\}$ and learning rate $\alpha \in \{1\mathrm{e}{-4}, 3\mathrm{e}{-4}, 1\mathrm{e}{-3}, 3\mathrm{e}{-3}\}$ on learning performance.

\begin{figure}[ht]
    \centering
    \includegraphics[width=0.8\linewidth]{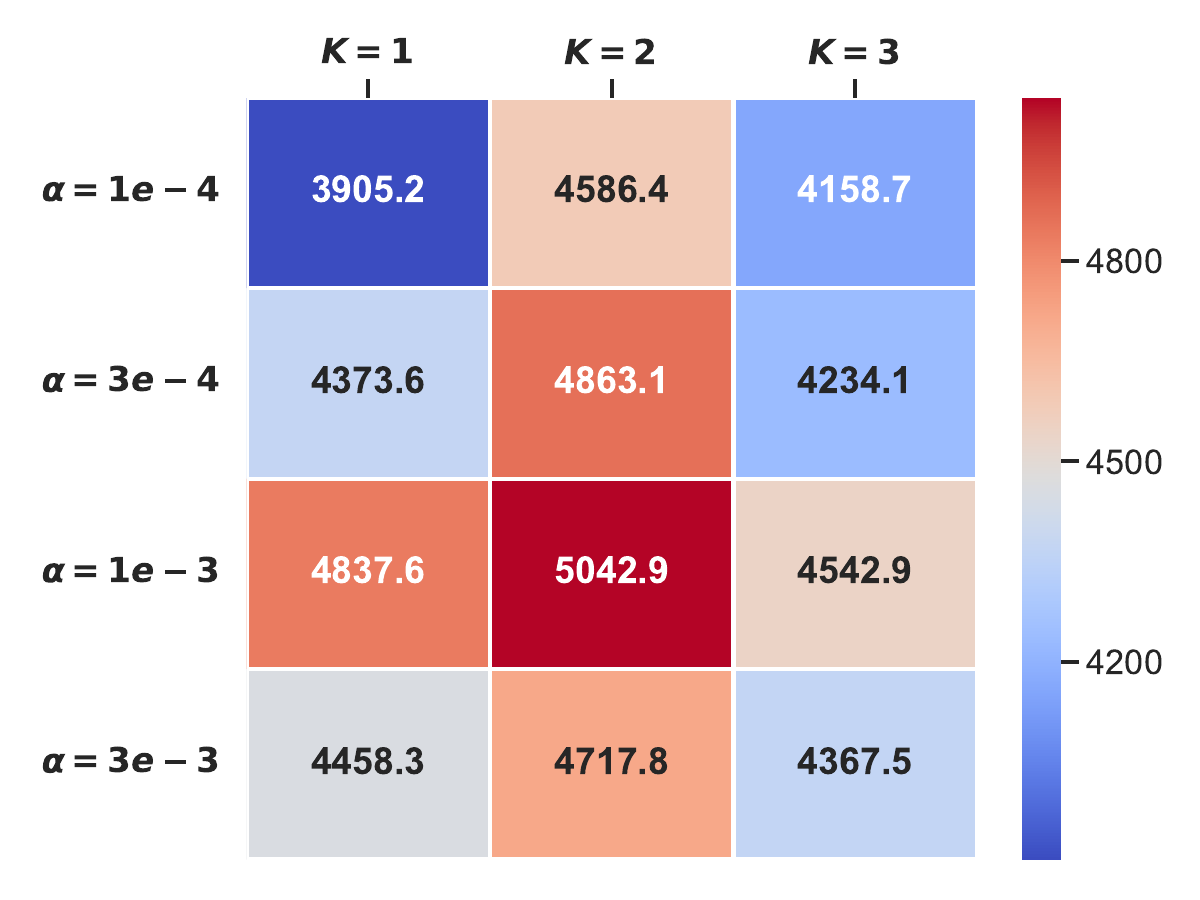}
    \caption{Hyperparameter analysis on Walker2d.}
    \label{fig:heatmap}
\end{figure}

As shown in Figure~\ref{fig:heatmap}, Reflex-SAC exhibits clear and consistent sensitivity patterns with respect to both the critic target update frequency $K$ and the learning rate $\alpha$. Performance peaks at moderate target update frequency ($K=2$) and an aggressive yet stable learning rate ($\alpha=1\mathrm{e}{-3}$), while overly frequent or infrequent target updates and excessively large learning rates lead to degraded performance. 

\begin{figure}[ht]
    \centering
    \begin{subfigure}{0.49\columnwidth}
        \centering
        \includegraphics[width=\linewidth]{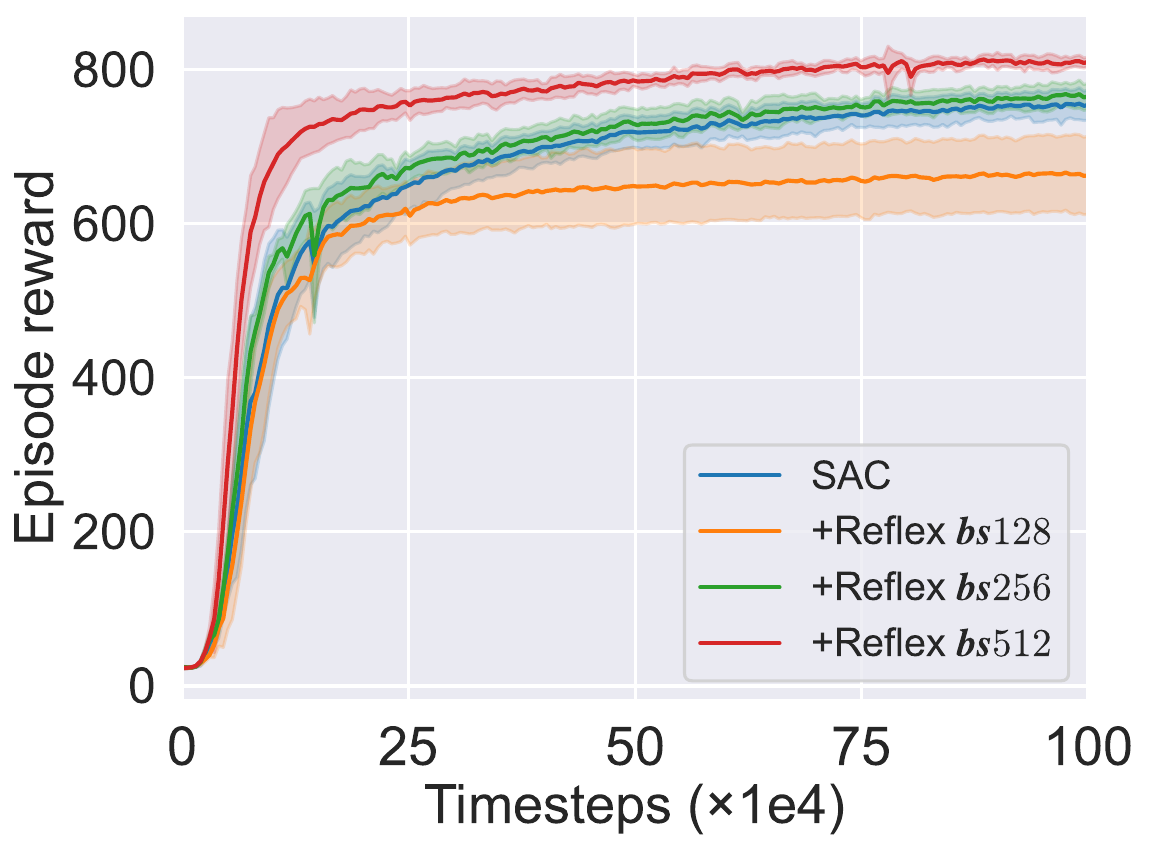}
        \caption{WalkerRun}
    \end{subfigure}
    \hfil
    \begin{subfigure}{0.49\columnwidth}
        \centering
        \includegraphics[width=\linewidth]{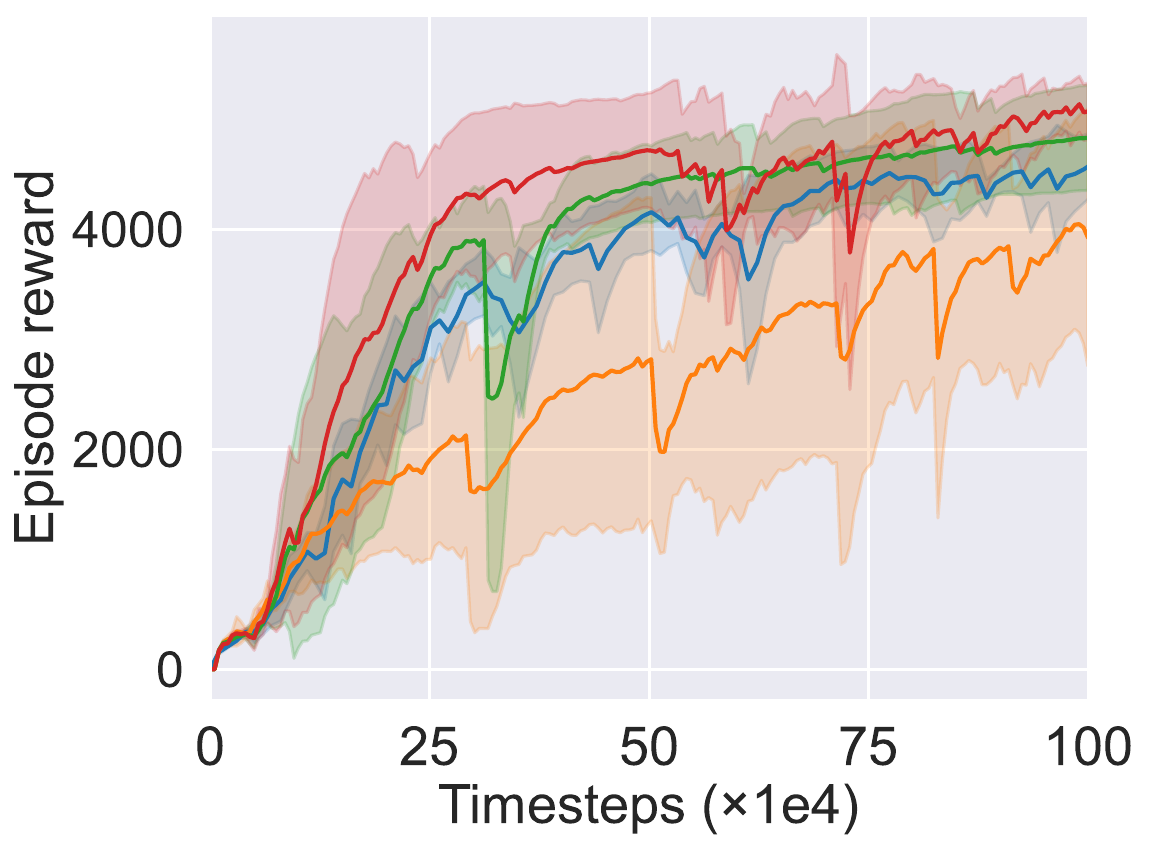}
        \caption{Walker2d}
    \end{subfigure}
    \caption{Analysis of batch\_size effect on Reflex-SAC.}
    \label{fig:batch_size}
\end{figure}

We additionally evaluate the impact of batch size with $K=2$ and $\alpha=1\mathrm{e}{-3}$ in Figure~\ref{fig:batch_size}. In contrast to DrQ in image-based RL, which achieves its best performance with $batch\_size = 512$ but remains relatively stable even under smaller batch sizes, state-based Reflex-SAC exhibits higher sensitivity to batch size.
All above results indicate that reflection symmetry regularization enables Reflex-SAC to remain well-performing under more aggressive learning settings, yielding improved robustness and sample efficiency.

\subsection{Applicability on TD3 Algorithm}
\label{sec:applicability}
\begin{figure}[ht]
    \centering
    \begin{subfigure}{0.49\columnwidth}
        \centering
        \includegraphics[width=\linewidth]{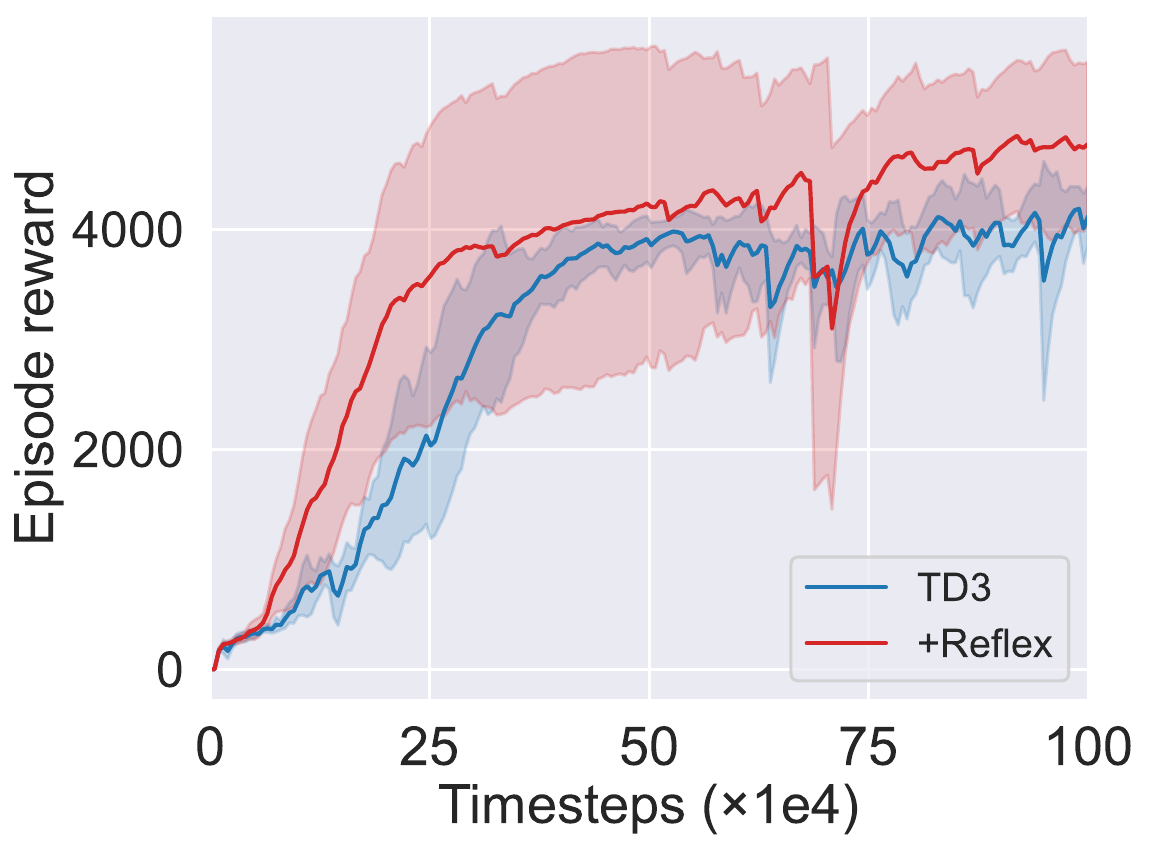}
        \caption{Walker2d}
    \end{subfigure}
    \hfil
    \begin{subfigure}{0.49\columnwidth}
        \centering
        \includegraphics[width=\linewidth]{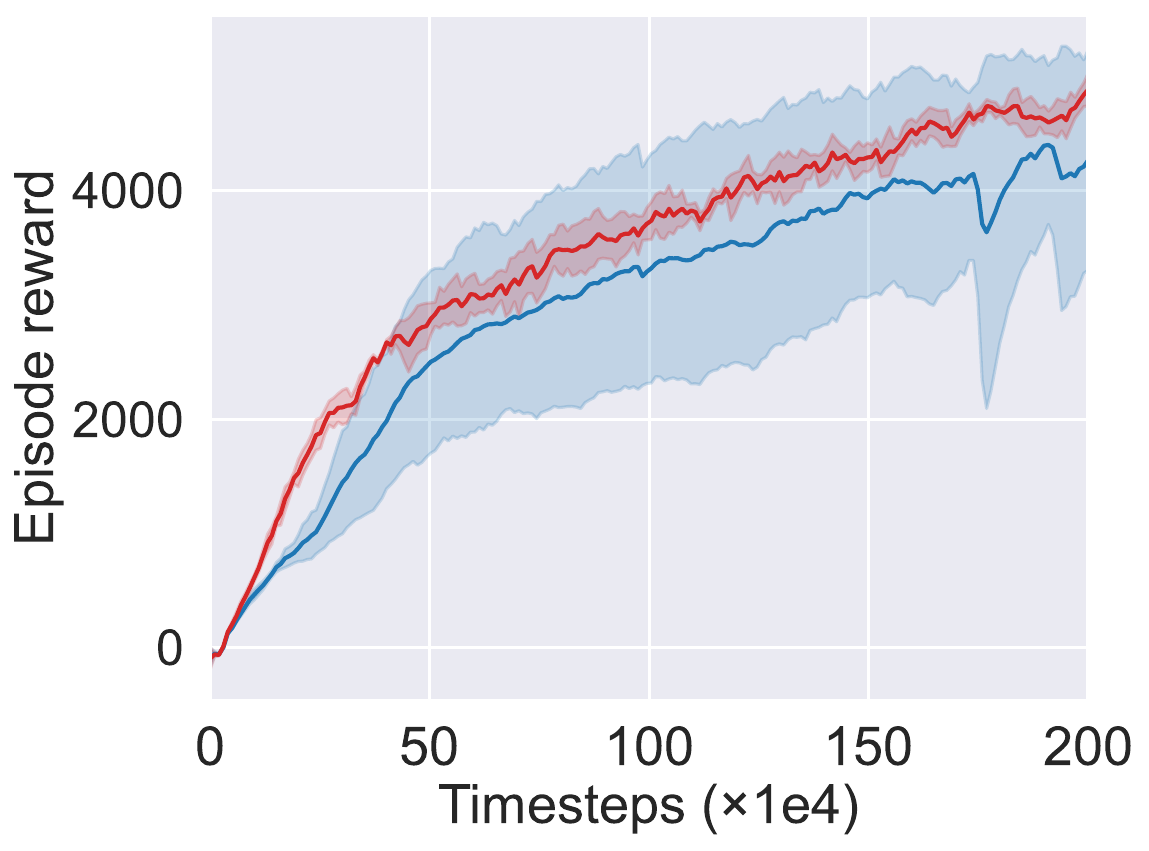}
        \caption{Ant}
    \end{subfigure}
    \caption{Applicability of Reflex on TD3 algorithm.}
    \label{fig:td3_result}
\end{figure}

To demonstrate the generality of Reflex, we further apply the proposed symmetry regularization term to TD3~\cite{fujimoto2018td3}, another widely used off-policy RL algorithm. 
We keep the philosophy consistent with Eqs.~\eqref{eq:sac_ref_target}\eqref{eq:sac_sym_q}.
As shown in Figure~\ref{fig:td3_result}, Reflex still performs favorably against the standard TD3 baseline, demonstrating its broad applicability across different RL algorithms.

\section{Conclusion}
In this work, we studied how reflection symmetry can be systematically exploited in state-based continuous control. We first formalized axial and bilateral reflection within the framework of group-invariant MDPs, and established that both optimal value functions and optimal policies admit equivariant solutions. 
Building on theoretical analysis, we proposed Reflex, a simple and general framework that integrates reflection symmetry into both on-policy and off-policy RL algorithms. Reflex leverages symmetry through regularization mechanisms tailored to different algorithmic settings, enabling effective information sharing across mirrored states and actions. Extensive experiments with PPO and SAC demonstrate that Reflex consistently improves sample efficiency and final performance across a range of continuous control tasks. And additional experiment on TD3 further validates the generality of Reflex.

\clearpage
\section*{Impact Statement}
This work proposes a symmetry-aware reinforcement learning framework that improves sample efficiency in state-based continuous control by systematically exploiting reflection symmetry. By integrating principled symmetry regularization into both on-policy and off-policy algorithms, the proposed approach can improve reinforcement learning algorithm performance while reducing redundant environment interactions. These improvements lower the computational and environmental costs associated with training reinforcement learning agents, particularly in robotics and real-world control applications. The method is general and does not introduce new safety risks beyond those commonly associated with existing reinforcement learning systems.

\bibliography{main}
\bibliographystyle{icml2026}

%%%%%%%%%%%%%%%%%%%%%%%%%%%%%%%%%%%%%%%%%%%%%%%%%%%%%%%%%%%%%%%%%%%%%%%%%%%%%%%
%%%%%%%%%%%%%%%%%%%%%%%%%%%%%%%%%%%%%%%%%%%%%%%%%%%%%%%%%%%%%%%%%%%%%%%%%%%%%%%
% APPENDIX
%%%%%%%%%%%%%%%%%%%%%%%%%%%%%%%%%%%%%%%%%%%%%%%%%%%%%%%%%%%%%%%%%%%%%%%%%%%%%%%
%%%%%%%%%%%%%%%%%%%%%%%%%%%%%%%%%%%%%%%%%%%%%%%%%%%%%%%%%%%%%%%%%%%%%%%%%%%%%%%
\newpage
\appendix
\onecolumn
\section{Reflection Transformation Rules}
\label{appendix:trans}
We consider a reflection with respect to an axis (or hyperplane) $\ell_u$ in $\mathbb{R}^d$. Let $u \in \mathbb{R}^d$ be a unit normal vector orthogonal to $\ell_u$, i.e., $\|u\|_2=1$. The reflection is the linear map that flips the component along $u$ while keeping the orthogonal component unchanged. For any $x\in\mathbb{R}^d$, decompose:
\begin{equation}
    x = x_{\parallel} + x_{\perp}, \qquad 
    x_{\parallel} := (u^\top x)u,\quad x_{\perp} := x - (u^\top x)u,
\end{equation}
and define $\tilde{x} := -x_{\parallel} + x_{\perp}$. This yields:
\begin{equation}
    \tilde{x} = x - 2(u^\top x)u = (I - 2uu^\top)x,
    \qquad\Rightarrow\qquad
    R := I - 2uu^\top .
\end{equation}
From $\|u\|_2=1$ we have:
\begin{equation}
    R^\top = R,\qquad
    R^2 = (I-2uu^\top)^2 = I,\qquad
    R^\top R = I,
\end{equation}
so $R$ is orthogonal and involutory. Moreover, $R$ is an improper orthogonal transform with $\det(R)=-1$.

\paragraph{Orientation.} Orientation is commonly expressed as either a unit quaternion $q \in \mathbb{R}^4$ or a rotation matrix $O \in \mathrm{SO(d)}$. Since quaternions and rotation matrices encode equivalent rotational information, we adopt rotation matrices for clarity. Let $O\in \mathrm{SO}(d)$ denote an orientation matrix that maps body-frame coordinates to world-frame coordinates. Under the reflected world coordinate system $v \mapsto Rv$ and reflected body coordinate system $w \mapsto Rw$, the same physical rotation must satisfy:
\begin{equation}
    R(Ow) = \tilde{O}(Rw) \Leftrightarrow \tilde{O} = R O R,  \quad \forall w.
\end{equation}

\paragraph{Euclidean vectors.}
For any Euclidean vector component $x_e\in\mathbb{R}^d$ (e.g., position, velocity, force) that transforms covariantly with the ambient coordinates, its reflected value follows directly from linearity:
\begin{equation}
    \tilde{x}_e = R x_e .
\end{equation}

\paragraph{Pseudovectors.}
Pseudovectors (e.g., angular velocity, torque) arise from the cross product / exterior product of Euclidean vectors and pick up an additional sign under improper orthogonal transformations. For $d=3$, let $a,b\in\mathbb{R}^3$ be Euclidean vectors and define the axial vector $x_p := a \times b$. Using the standard identity for orthogonal transforms $A\in O(3)$:
\begin{equation}
    (Aa)\times(Ab) = \det(A)\,A(a\times b),
\end{equation}
and taking $A=R$ with $\det(R)=-1$, we obtain:
\begin{equation}
    \widetilde{x}_p
    = (Ra)\times(Rb)
    = \det(R)\,R(a\times b)
    = -R x_p .
\end{equation}

\paragraph{Angular (phase) variables.}
For an angle variable $x_\theta$ represented on the unit circle by:
\begin{equation}
    v(x_\theta) := 
    \begin{bmatrix}
    \cos x_\theta \\ \sin x_\theta
    \end{bmatrix}\in \mathbb{R}^2,
\end{equation}
its reflection is induced by the 2D reflection matrix $R\in\mathbb{R}^{2\times 2}$:
\begin{equation}
    \widetilde{v} = R\,v(x_\theta).
\end{equation}
Mapping back to an angle via $\mathrm{atan2}$ gives:
\begin{equation}
    \tilde{x}_\theta = \mathrm{atan2}((R v(x_\theta))_2,\,(R v(x_\theta))_1),
\end{equation}
which we abbreviate as $\tilde{x}_\theta = \mathrm{atan2}(R x_\theta)$ when $x_\theta$ is stored in its $(\cos,\sin)$ embedding.

\paragraph{Invariant scalars.}
For invariant scalar quantities $x_q$ (e.g., time, temperature), they remain unchanged under reflection:
\begin{equation}
    \tilde{x}_q = x_q .
\end{equation}

\section{Proof}
\subsection{Proof of Lemma~\ref{lemma:bellman}}
\label{app:proof_bellman}
\begin{proof}
For any state $s \in \mathcal{S}$, by the definition of the Bellman optimality operator:
\begin{equation}
    (\mathcal{T}(V \circ g))(s) = \max_{a \in \mathcal{A}} \left[ r(s, a) + \gamma \mathbb{E}_{s' \sim \mathcal{P}(\cdot|s, a)} [V(gs')] \right].
\end{equation}
According to the properties of a $G$-invariant MDP, we have $r(s, a) = r(gs, ga)$ and $\mathcal{P}(s'|s, a) = \mathcal{P}(gs'|gs, ga)$. Substituting these into the equation:
\begin{equation}
    (\mathcal{T}(V \circ g))(s) = \max_{a \in \mathcal{A}} \left[ r(gs, ga) + \gamma \sum_{s' \in \mathcal{S}} \mathcal{P}(gs' | gs, ga) V(gs') \right].
\end{equation}
Let $s'' = gs'$. Since the group action $g$ is a bijection on the state space $\mathcal{S}$, we can rewrite the summation over $s''$:
\begin{equation}
    (\mathcal{T}(V \circ g))(s) = \max_{a \in \mathcal{A}} \left[ r(gs, ga) + \gamma \sum_{s'' \in \mathcal{S}} \mathcal{P}(s'' | gs, ga) V(s'') \right].
\end{equation}
Since $g$ is also a bijection on the action space $\mathcal{A}$, the maximization over $a \in \mathcal{A}$ is equivalent to the maximization over $a' = ga \in \mathcal{A}$:
\begin{equation}
    (\mathcal{T}(V \circ g))(s) = \max_{a' \in \mathcal{A}} \left[ r(gs, a') + \gamma \mathbb{E}_{s'' \sim \mathcal{P}(\cdot|gs, a')} [V(s'')] \right].
\end{equation}
The right-hand side is exactly the definition of the Bellman optimality operator evaluated at state $gs$:
\begin{equation}
    (\mathcal{T}(V \circ g))(s) = (\mathcal{T}V)(gs) = ((\mathcal{T}V) \circ g)(s).
\end{equation}
Since $s$ is arbitrary, the result $\mathcal{T}(V \circ g) = (\mathcal{T}V) \circ g$ holds.
\end{proof}

\subsection{Proof of Lemma~\ref{lemma:value_invariance}}
\label{app:value_invariance}
\begin{proof}
The proof is divided into two parts: the invariance of $V^*$ and the invariance of $Q^*$.

\textbf{Part 1: Invariance of $V^*$.}
The optimal state-value function $V^*$ is the unique fixed point of the Bellman optimality operator $\mathcal{T}$, such that $\mathcal{T}V^* = V^*$. For any $g \in G$, we consider the function $V^* \circ g$. Applying the Bellman optimality operator $\mathcal{T}$ to $V^* \circ g$:
\begin{equation}
    \mathcal{T}(V^* \circ g) = (\mathcal{T}V^*) \circ g,
\end{equation}
where the equality follows directly from the equivariance of the Bellman operator established in Lemma~\ref{lemma:bellman}. Since $V^*$ is the fixed point ($\mathcal{T}V^* = V^*$), we have:
\begin{equation}
    \mathcal{T}(V^* \circ g) = V^* \circ g.
\end{equation}
This implies that $V^* \circ g$ is also a fixed point of the operator $\mathcal{T}$. By the contraction mapping theorem, the Bellman optimality operator has a unique fixed point. Therefore, it must hold that:
\begin{equation}
    V^* = V^* \circ g, \quad \forall g \in G,
\end{equation}
which proves $V^*(s) = V^*(gs)$ for all $s \in \mathcal{S}$.

\textbf{Part 2: Invariance of $Q^*$.}
By definition, the optimal action-value function is given by:
\begin{equation}
    Q^*(s, a) = r(s, a) + \gamma \mathbb{E}_{s' \sim \mathcal{P}(\cdot|s, a)} [V^*(s')].
\end{equation}
Evaluating $Q^*$ at the transformed state-action pair $(gs, ga)$:
\begin{equation}
    Q^*(gs, ga) = r(gs, ga) + \gamma \mathbb{E}_{s'' \sim \mathcal{P}(\cdot|gs, ga)} [V^*(s'')].
\end{equation}
According to the properties of a $G$-invariant MDP, we substitute $r(gs, ga) = r(s, a)$ and $\mathcal{P}(s'' | gs, ga) = \mathcal{P}(g^{-1}s'' | s, a)$:
\begin{equation}
    Q^*(gs, ga) = r(s, a) + \gamma \sum_{s'' \in \mathcal{S}} \mathcal{P}(g^{-1}s'' | s, a) V^*(s'').
\end{equation}
We also have $\mathcal{P}(s'' | gs, ga) = \mathcal{P}(gs' | gs, ga)$, which implies $s'' = gs'$, we perform a change of variable by setting $s' = g^{-1}s''$. Since $g$ is a bijection on $\mathcal{S}$, we rewrite the sum:
\begin{equation}
    Q^*(gs, ga) = r(s, a) + \gamma \sum_{s' \in \mathcal{S}} \mathcal{P}(s' | s, a) V^*(gs').
\end{equation}
From the invariance of $V^*$ proved in Part 1, we have $V^*(gs') = V^*(s')$. Substituting this into the equation:
\begin{equation}
    Q^*(gs, ga) = r(s, a) + \gamma \sum_{s' \in \mathcal{S}} \mathcal{P}(s' | s, a) V^*(s') = Q^*(s, a).
\end{equation}
Thus, $Q^*(s, a) = Q^*(gs, ga)$ for all $g \in G, s \in \mathcal{S}, a \in \mathcal{A}$.
\end{proof}

\subsection{Proof of Theorem~\ref{theorem:policy_equi}}
\label{app:policy_equi}

\begin{proof}
The proof leverages the invariance of the optimal action-value function $Q^*$ to show that the set of optimal actions transforms equivariantly under the group action $G$.

\textbf{Part 1: Equivariance of the Optimal Action Set.}
An action $a$ belongs to the optimal set $\mathcal{A}^*(s)$ if it maximizes $Q^*(s, \cdot)$:
\begin{equation}
    a \in \mathcal{A}^*(s) \iff Q^*(s, a) \geq Q^*(s, a'), \quad \forall a' \in \mathcal{A}.
\end{equation}
From Lemma~\ref{lemma:value_invariance}, we have the invariance $Q^*(s, a) = Q^*(gs, ga)$. Substituting this into the inequality:
\begin{equation}
    Q^*(gs, ga) \geq Q^*(gs, ga'), \quad \forall a' \in \mathcal{A}.
\end{equation}
Since the group action $g$ is a bijection on the action space $\mathcal{A}$, as $a'$ ranges over all possible actions in $\mathcal{A}$, $ga'$ also covers the entire set $\mathcal{A}$. Thus, by a change of variables $\bar{a} = g a'$, we can rewrite the condition as:
\begin{equation}
    Q^*(gs, ga) \geq Q^*(gs, \bar{a}), \quad \forall \bar{a} \in \mathcal{A}.
\end{equation}
This is precisely the definition of $ga$ being an optimal action at state $gs$, i.e., $ga \in \mathcal{A}^*(gs)$. Consequently, the sets transform as $\mathcal{A}^*(gs) = g\mathcal{A}^*(s)$.

\textbf{Part 2: Deterministic Policy.} We construct a deterministic optimal policy that is equivariant under the group action $G$.
Since $\mathcal{A}^*(gs) = g\mathcal{A}^*(s)$ from Part 1, we can construct an equivariant policy by consistently selecting corresponding actions across symmetric states. Specifically, fix an arbitrary ordering $\prec$ on the action space $\mathcal{A}$. Define the deterministic policy:
\begin{equation}
\mu^*(s) = \min_{\prec} \mathcal{A}^*(s).
\end{equation}

For any $g \in G$ and $s \in \mathcal{S}$, we have:
\begin{align}
\mu^*(gs) &= \min_{\prec} \mathcal{A}^*(gs) \\
&= \min_{\prec} g\mathcal{A}^*(s) \quad \text{(by Part 1)} \\
&= g \cdot \min_{\prec} \mathcal{A}^*(s) \quad \text{(since $g$ preserves the ordering)} \\
&= g\mu^*(s).
\end{align}

This satisfies the definition of an equivariant mapping:
\begin{equation}
\mu^*(gs) = g\mu^*(s), \quad \forall g \in G, s \in \mathcal{S}. \tag{30}
\end{equation}

\textbf{Part 3: Stochastic Policy.}
For a stochastic policy $\pi^*$, we define $\pi^*$ as a uniform distribution over the optimal action set $\mathcal{A}^*(s)$:
\begin{equation}
    \pi^*(a|s) = \begin{cases} \frac{1}{|\mathcal{A}^*(s)|}, & \text{if } a \in \mathcal{A}^*(s) \\ 0, & \text{otherwise} \end{cases}
\end{equation}
To verify the condition $\pi^*(a|gs) = \pi^*(g^{-1}a|s)$, consider two cases:
If $a \in \mathcal{A}^*(gs)$, then $g^{-1}a \in \mathcal{A}^*(s)$ from Part 1. Since $g$ is a bijection, the cardinality is preserved: $|\mathcal{A}^*(gs)| = |g\mathcal{A}^*(s)| = |\mathcal{A}^*(s)|$. Thus:
\begin{equation}
    \pi^*(a|gs) = \frac{1}{|\mathcal{A}^*(gs)|} = \frac{1}{|\mathcal{A}^*(s)|} = \pi^*(g^{-1}a|s).
\end{equation}
If $a \notin \mathcal{A}^*(gs)$, then $g^{-1}a \notin \mathcal{A}^*(s)$, both sides of the equation equal zero. To verify optimality, note that:
\begin{equation}
V^{\pi^*}(s) = \sum_{a \in \mathcal{A}^*(s)} \frac{1}{|\mathcal{A}^*(s)|} Q^*(s,a) = \sum_{a \in \mathcal{A}^*(s)} \frac{1}{|\mathcal{A}^*(s)|} V^*(s) = V^*(s),
\end{equation}
where we used the fact that $Q^*(s,a) = V^*(s)$ for all $a \in \mathcal{A}^*(s)$ by definition of the optimal action set. Since $V^{\pi^*} = V^*$, the policy $\pi^*$ is optimal. Thus, an equivariant stochastic optimal policy always exists in $\mathcal{M}_G$.
\end{proof}

\section{Algorithms.}
\label{app:algorithms}

\begin{algorithm}[ht]
\caption{Reflex-PPO}
\label{alg:ppo}
\begin{algorithmic}[1]
\STATE Initialize policy parameters $\theta$, value function parameters $\phi$, batch size $B$, number of epochs $K$, horizon $H$
\FOR{iteration $k = 0, 1, 2, \dots$}
    \STATE Run $\pi_{\theta}$ for $H$ timesteps, get $\tau = \{(s_t, a_t, r_t)\}$
    \FOR{epoch $= 1$ to $K$}
        \STATE Sample mini-batch $\mathcal{B} = \{(s_i, a_i, r_i)\}$ of size $B$ from $\tau$
        \STATE \textcolor{red}{Construct reflected mini-batch $\tilde{\mathcal{B}} = \{(\tilde{s}_i, \tilde{a}_i, r_i)\}$ by transformation rule in Table~\ref{tab:reflection_rules} and Eq.~\eqref{eq:bilateral_reflection}}
        \STATE Update policy parameters $\theta$ by Eq.~\eqref{eq:reflex-ppo_Aloss}\eqref{eq:ppo_Aloss}
        \STATE Update value function parameters $\phi$ by Eq.~\eqref{eq:reflex-ppo_Closs}\eqref{eq:ppo_Closs}
    \ENDFOR
\ENDFOR
\end{algorithmic}
\end{algorithm}

\begin{algorithm}[ht]
\caption{Reflex-SAC}
\label{alg:sac}
\begin{algorithmic}[1]
\STATE Initialize empty replay buffer $\mathcal{R}$, Q-function parameters $\phi_1$, $\phi_2$, policy parameters $\theta$, and temperature $\alpha$, target Q-function parameters $\bar{\phi}_i \leftarrow \phi_i$ for $i \in \{1,2\}$, soft target update coefficient $\tau$, total timesteps $T$
\FOR{each timestep $t = 1, \dots, T$}
    \STATE Collect transition $(s_t, a_t, r_t, s_{t+1})$ and store in replay buffer $\mathcal{R}$
    \STATE Sample a batch of transitions $\mathcal{B}$ from $\mathcal{R}$
    \STATE \textcolor{red}{Construct reflected batch $\tilde{\mathcal{B}} = \{(\tilde{s}_i, \tilde{a}_i, r_i)\}$ by transformation rule in Table~\ref{tab:reflection_rules} and Eq.~\eqref{eq:bilateral_reflection}}
    \STATE Compute target value $y$ and \textcolor{red}{$\tilde{y}$} by Eq.~\eqref{eq:sac_target}
    \STATE \textcolor{red}{Compute final target value $\bar{y}$ by Eq.~\eqref{eq:sac_ref_target}}       
    \STATE Update Q-function parameters $\phi$ by Eq.~\eqref{eq:sac_sym_q}

    \IF{$t \bmod \texttt{policy\_frequency} = 0$}
        \STATE Update policy parameters $\theta$ by $\mathcal{L}_{\pi}^\theta = - \mathbb{E}_{a \sim \pi} \Big[ Q_\phi(s_t,a) - \alpha \log \pi_\theta(a|s_t) \Big].$
        \STATE Adjust temperature $\alpha$
    \ENDIF
        
    \IF{$t \bmod \texttt{target\_network\_frequency} = 0$}
        \STATE Update target networks $\bar{\phi}_i \leftarrow \tau \phi_i + (1-\tau)\bar{\phi}_i$
    \ENDIF
    
\ENDFOR
\end{algorithmic}
\end{algorithm}

\clearpage 
\section{Experimental Details}
\subsection{Environments and Tasks}
\label{app:env}
\begin{itemize}
    \item \textbf{CartpoleBalance}: This environment originates from control theory and builds on the cartpole environment based on the~\citet{barto2012neuronlike}.    The cartpole is a classic control problem where a pole is attached to a cart that can move along a one-dimensional track. The goal is to balance the pole in an upright position by applying forces to the cart.
    \item \textbf{CartpoleThreePoles}: This environment extends the classic cartpole task by attaching three poles serially to a single cart. The cart can move along a one-dimensional track, and control forces are applied to the cart. The objective is to balance all three poles in an upright configuration simultaneously 
    \item \textbf{InvertedDoublePendulum}: This environment consists of a cart that can be moved linearly, with a pole attached to one end and having another end free. The cart can be pushed left or right, and the goal is to balance the pole on top of the cart by applying forces to the cart.
    \item \textbf{Walker2d}: The walker is a two-dimensional bipedal robot consisting of seven main body parts - a single torso at the top (with the two legs splitting after the torso), two thighs in the middle below the torso, two legs below the thighs, and two feet attached to the legs on which the entire body rests. The goal is to walk in the forward (right) direction by applying torque to the six hinges connecting the seven body parts.
    \item \textbf{WalkerRun}: An improved planar walker based on the one introduced in \citet{lillicrap2015ddpg}, which models a two-dimensional bipedal robot with articulated legs. Compared to the standard walking task, the run variant includes an additional reward component that explicitly encourages higher forward velocity. The agent must coordinate joint torques to maintain balance while achieving faster locomotion in the forward (right) direction.
    \item \textbf{Ant}: This environment is based on the one introduced by~\citet{schulman2015high}. The ant is a 3D quadruped robot consisting of a torso (free rotational body) with four legs attached to it, where each leg has two body parts. The goal is to coordinate the four legs to move in the forward (right) direction by applying torque to the eight hinges connecting the two body parts of each leg and the torso (nine body parts and eight hinges).
\end{itemize}

\begin{figure}[!h]
    \centering

    \begin{subfigure}{0.33\textwidth}
        \centering
        \includegraphics[width=\linewidth]{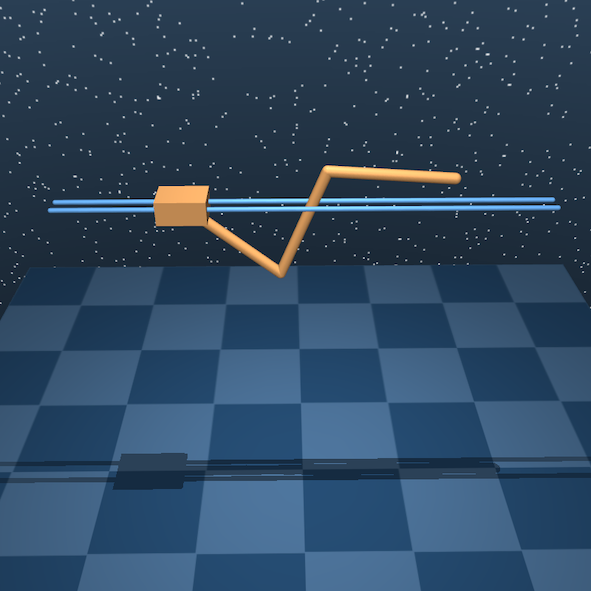}
        \caption{Cartpole}
    \end{subfigure}
    \hfill
    \begin{subfigure}{0.33\textwidth}
        \centering
        \includegraphics[width=\linewidth]{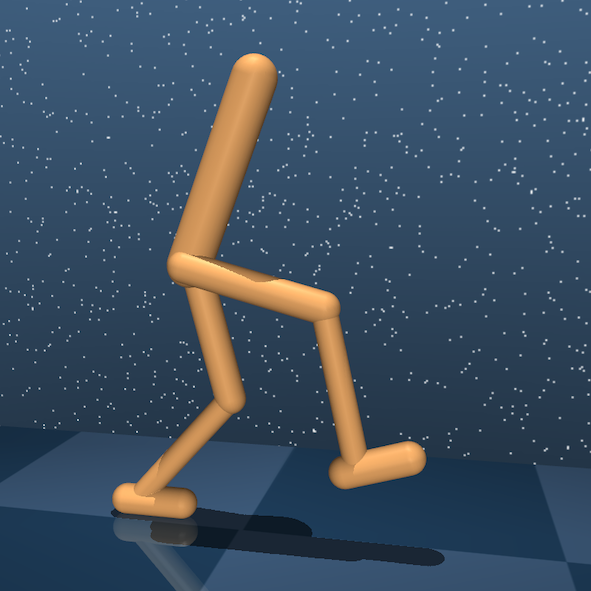}
        \caption{Walker}
    \end{subfigure}
    \hfill
    \begin{subfigure}{0.33\textwidth}
        \centering
        \includegraphics[width=\linewidth]{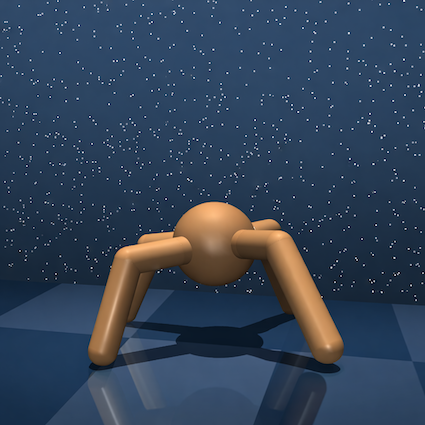}
        \caption{Ant}
    \end{subfigure}

    \caption{Illustrations of representative tasks used in our experiments. \textbf{(a)} The cart-pole illustration applies to the CartpoleBalance, CartpoleThreePoles, and InvertedDoublePendulum tasks. They only differ from different pole configurations and environments source. \textbf{(b)} The walker illustration represents both Walker2d and WalkerRun tasks. \textbf{(c)} The quadruped illustration corresponds to the Ant tasks.}
\end{figure}

\clearpage
\subsection{Hyperparameters}
\label{app:hyperparameters}

\begin{table}[ht]
    \centering
    \caption{PPO hyperparameters.}
    \begin{tabular}{l|c}
        \toprule
        \textbf{Hyperparameter} & \textbf{Value} \\
        \midrule
        Horizon (T) & 2048 \\
        Hidden layer size & 64\\
        Optimizer & Adam~\cite{kingma2014adam} \\
        Actor learning rate & $3\times10^{-4}$ \\
        Critic learning rate & $3\times10^{-4}$ \\
        Num. epochs & 10 \\
        Minibatch size & 64 \\
        Clip Range ($\mathcal{E}$) & 0.2 \\
        Discount ($\gamma$) & 0.99 \\
        GAE parameter ($\lambda$) & 0.95 \\
        \bottomrule
    \end{tabular}
\end{table}

\begin{table}[ht]
    \centering
    \caption{SAC hyperparameters, $\dagger$ denotes values used in Reflex-SAC.}
    \begin{tabular}{l|c}
        \toprule
        \textbf{Hyperparameter} & \textbf{Value} \\
        \midrule
        Optimizer & Adam~\cite{kingma2014adam} \\
        Actor learning rate & $3\times10^{-4}$ ($1\times10^{-3}\dagger$) \\
        Critic learning rate & $1\times10^{-3}$ \\
        Batch size & 256 ($512^\dagger$)\\
        Replay buffer size & $10^6$ \\
        Hidden layer size & 256\\
        Discount ($\gamma$) & 0.99 \\
        Warmup steps & 5000 \\
        Target smoothing coefficient ($\tau$) & 0.005 \\
        Target update interval & 1 ($2^\dagger$)\\
        Policy update interval & 2\\
        Alpha ($\alpha$) & 0.2 \\
        Autotune alpha & true \\
        \bottomrule
    \end{tabular}
\end{table}

\clearpage
\subsection{Complete Main Results}
\label{app:full_results}
\begin{figure}[!h]
    \centering

    \begin{subfigure}{0.33\textwidth}
        \centering
        \includegraphics[width=\linewidth]{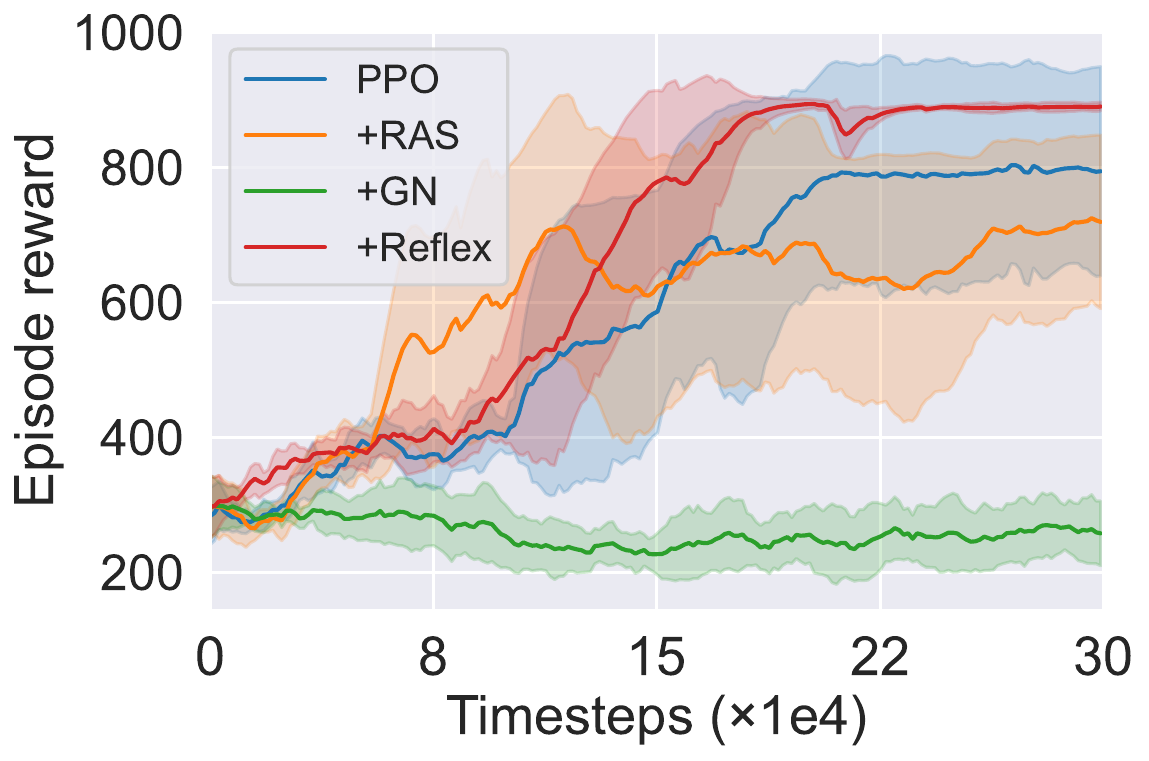}
        \caption{CartpoleBalance}
    \end{subfigure}
    \hfill
    \begin{subfigure}{0.33\textwidth}
        \centering
        \includegraphics[width=\linewidth]{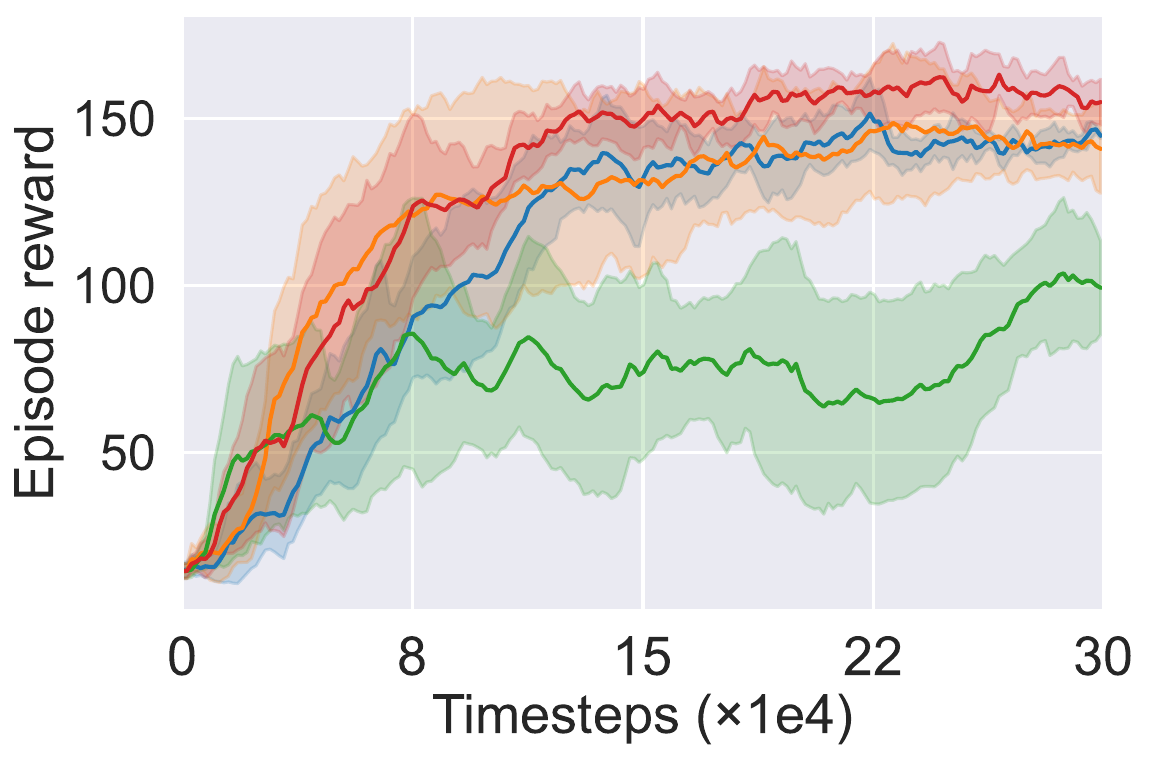}
        \caption{CartpoleThreePoles}
    \end{subfigure}
    \hfill
    \begin{subfigure}{0.33\textwidth}
        \centering
        \includegraphics[width=\linewidth]{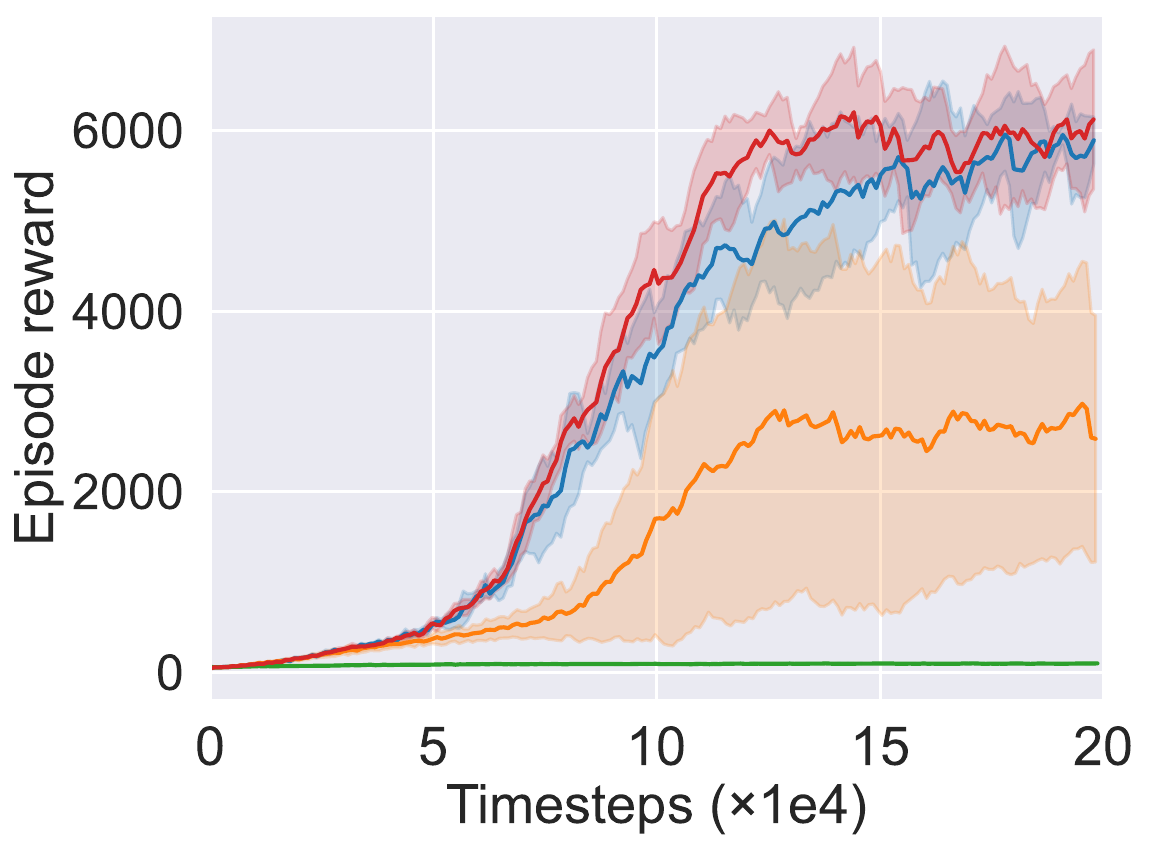}
        \caption{InvertedDoublePendulum}
    \end{subfigure}

    \begin{subfigure}{0.33\textwidth}
        \centering
        \includegraphics[width=\linewidth]{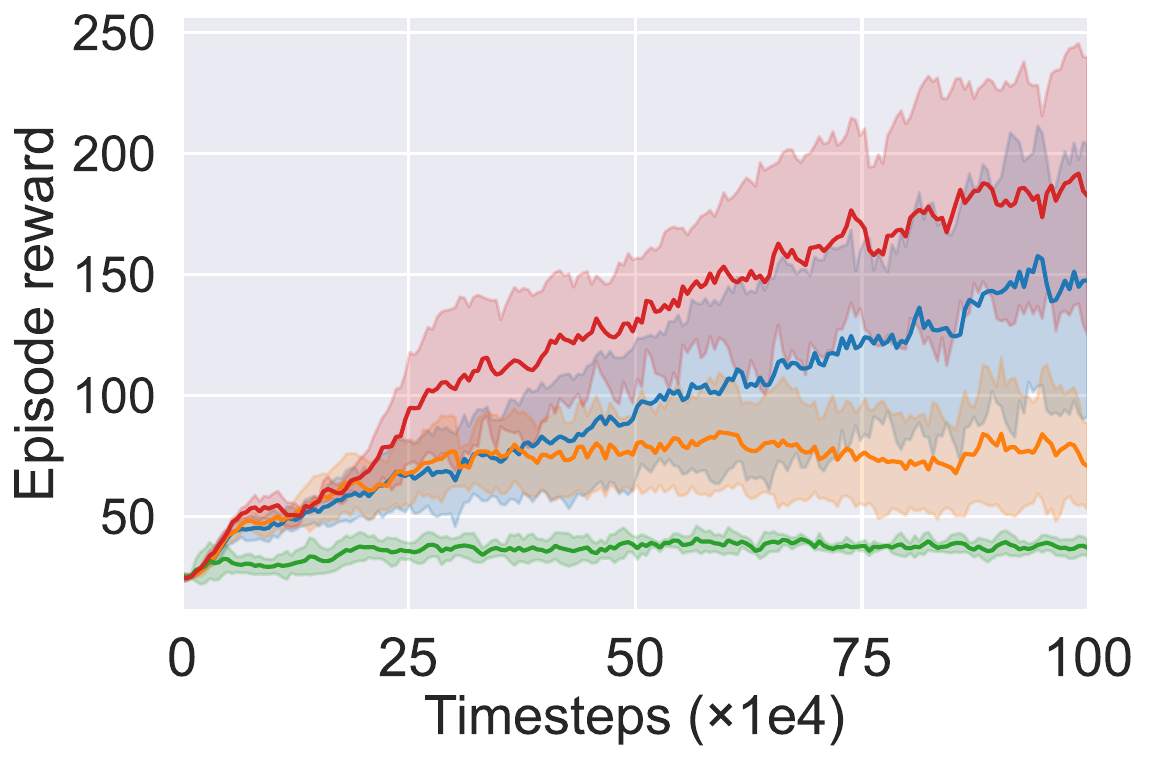}
        \caption{WalkerRun}
    \end{subfigure}
    \hfill
    \begin{subfigure}{0.33\textwidth}
        \centering
        \includegraphics[width=\linewidth]{figures/PPO/Walker2d-v4.pdf}
        \caption{Walker2d}
    \end{subfigure}
    \hfill
    \begin{subfigure}{0.33\textwidth}
        \centering
        \includegraphics[width=\linewidth]{figures/PPO/Ant-v4.pdf}
        \caption{Ant}
    \end{subfigure}

    \caption{Learning curves of PPO-based methods on all evaluated tasks.}
\end{figure}

\begin{figure}[!h]
    \centering

    \begin{subfigure}{0.33\textwidth}
        \centering
        \includegraphics[width=\linewidth]{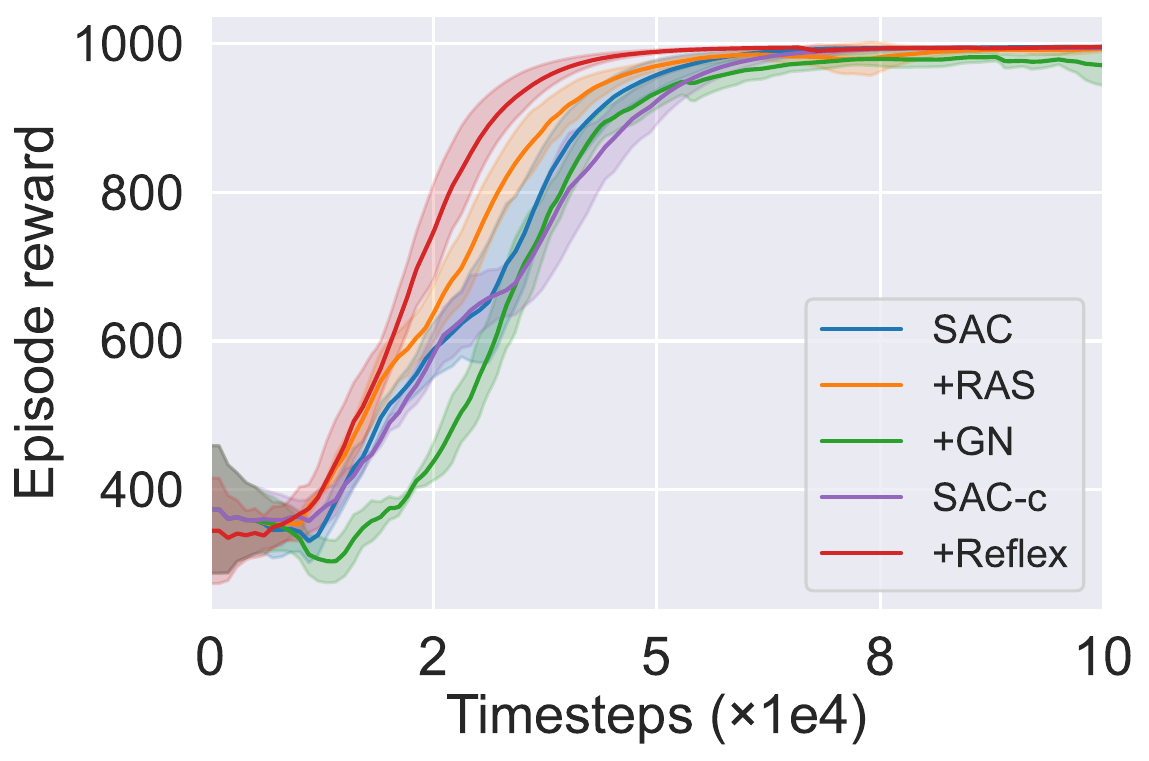}
        \caption{CartpoleBalance}
    \end{subfigure}
    \hfill
    \begin{subfigure}{0.33\textwidth}
        \centering
        \includegraphics[width=\linewidth]{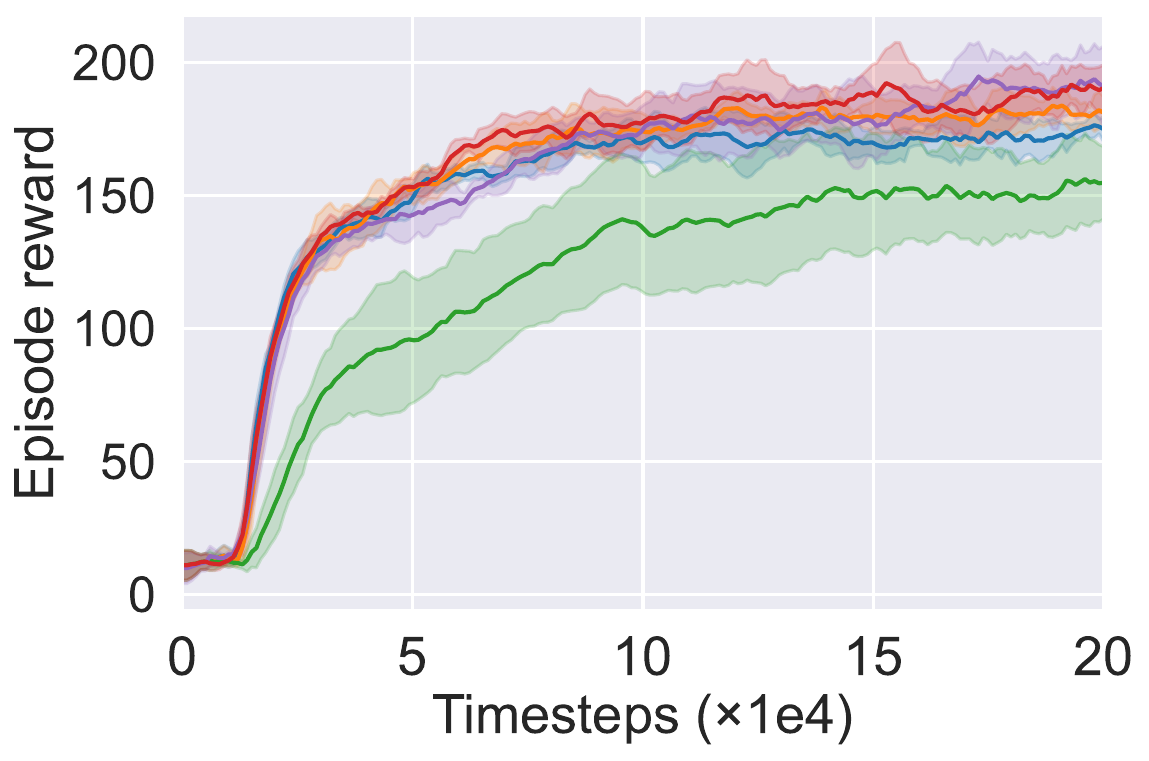}
        \caption{CartpoleThreePoles}
    \end{subfigure}
    \hfill
    \begin{subfigure}{0.33\textwidth}
        \centering
        \includegraphics[width=\linewidth]{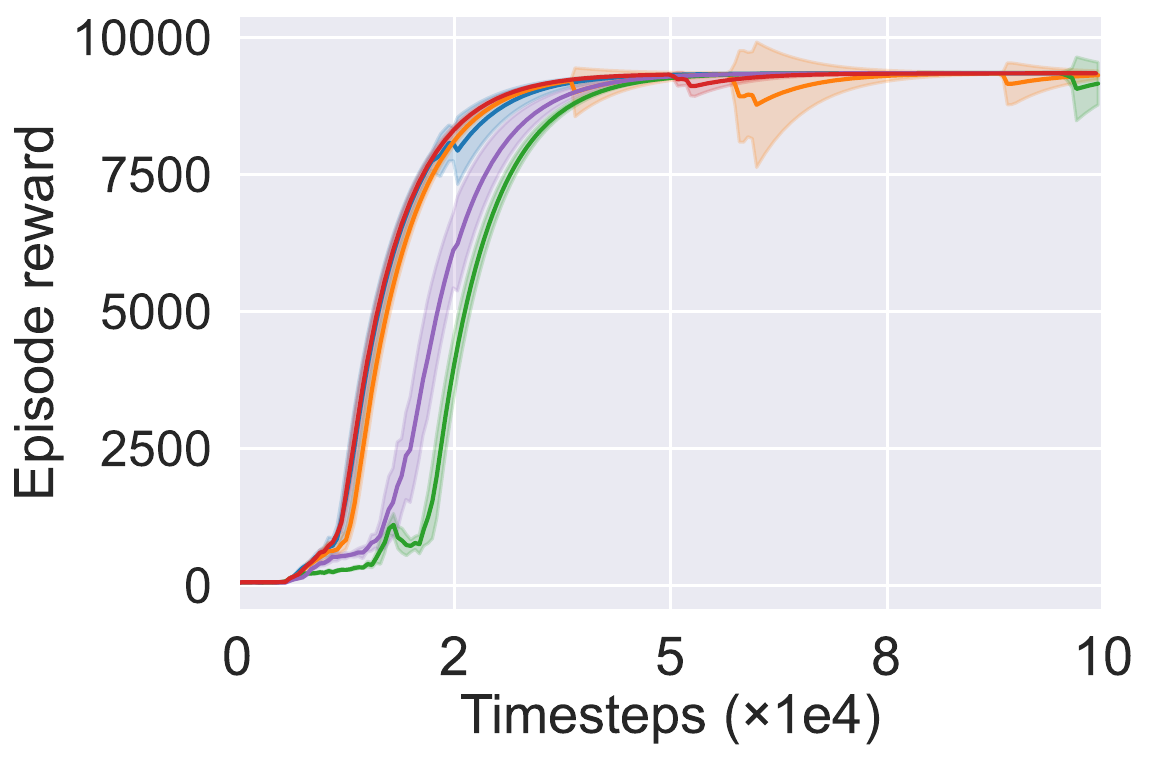}
        \caption{InvertedDoublePendulum}
    \end{subfigure}

    \begin{subfigure}{0.33\textwidth}
        \centering
        \includegraphics[width=\linewidth]{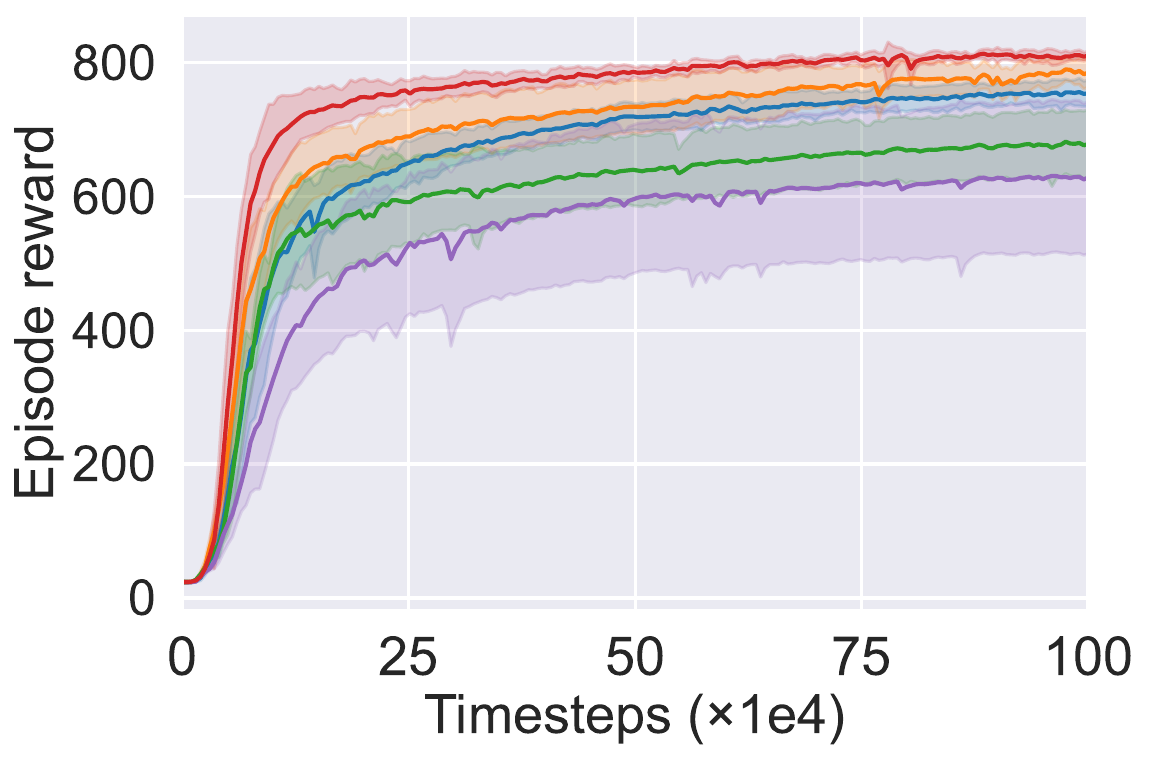}
        \caption{WalkerRun}
    \end{subfigure}
    \hfill
    \begin{subfigure}{0.33\textwidth}
        \centering
        \includegraphics[width=\linewidth]{figures/SAC/Walker2d-v4.pdf}
        \caption{Walker2d}
    \end{subfigure}
    \hfill
    \begin{subfigure}{0.33\textwidth}
        \centering
        \includegraphics[width=\linewidth]{figures/SAC/Ant-v4.pdf}
        \caption{Ant}
    \end{subfigure}

    \caption{Learning curves of SAC-based methods on all evaluated tasks.}
\end{figure}

\clearpage
\subsection{Complete Results on $w$ Analysis}
\begin{figure}[!h]
    \centering

    \begin{subfigure}{0.33\textwidth}
        \centering
        \includegraphics[width=\linewidth]{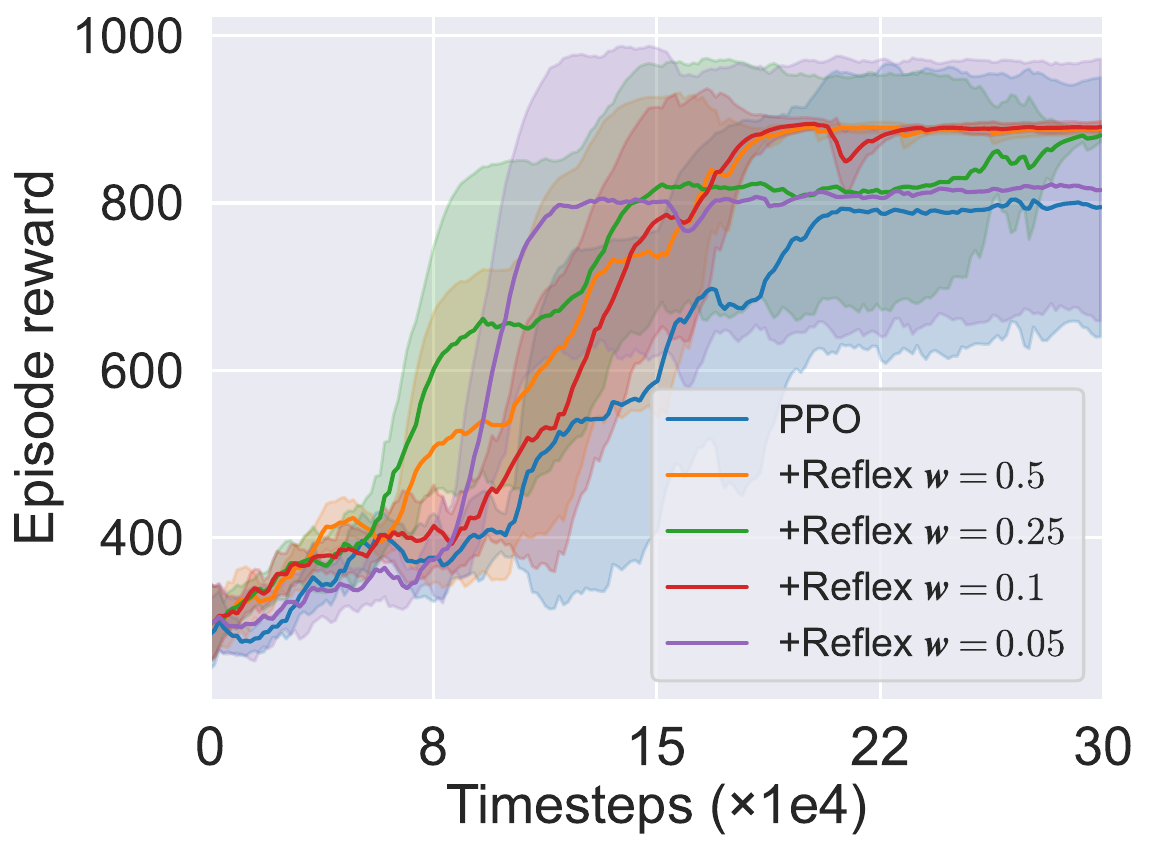}
        \caption{CartpoleBalance}
    \end{subfigure}
    \hfill
    \begin{subfigure}{0.33\textwidth}
        \centering
        \includegraphics[width=\linewidth]{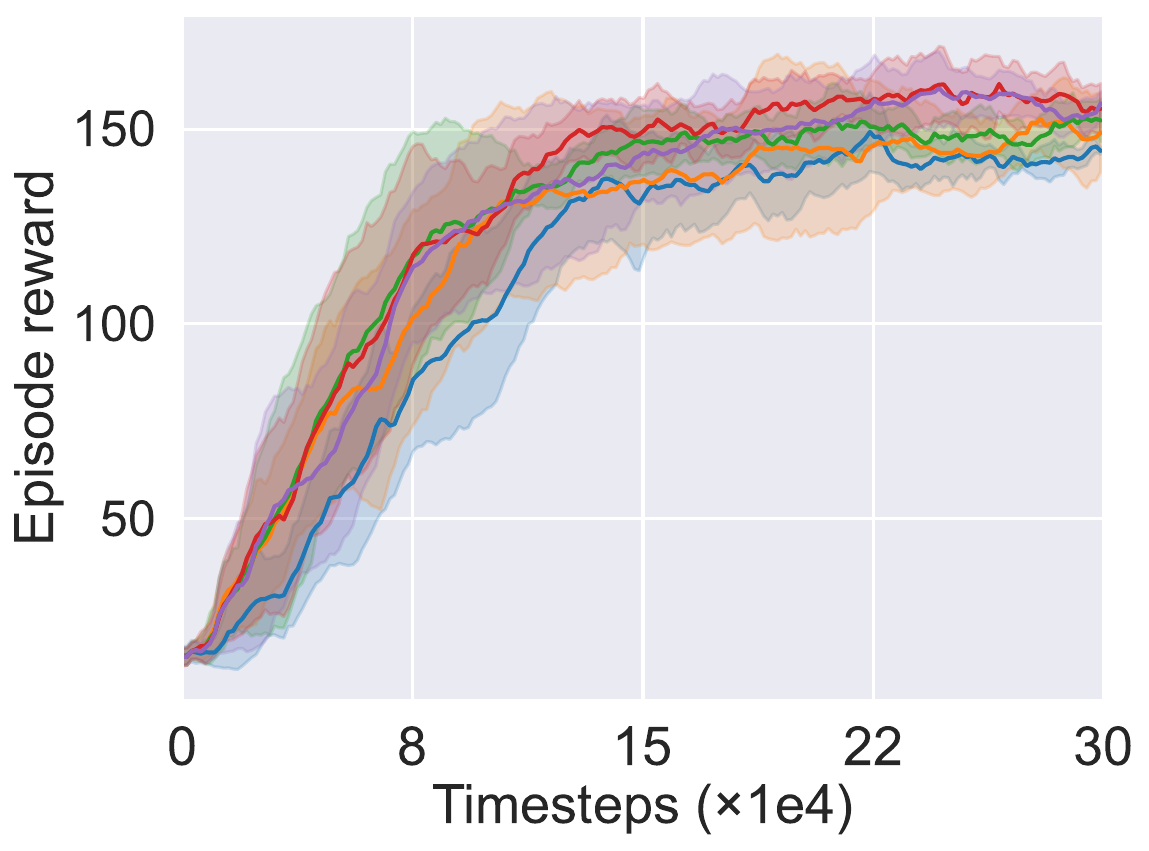}
        \caption{CartpoleThreePoles}
    \end{subfigure}
    \hfill
    \begin{subfigure}{0.33\textwidth}
        \centering
        \includegraphics[width=\linewidth]{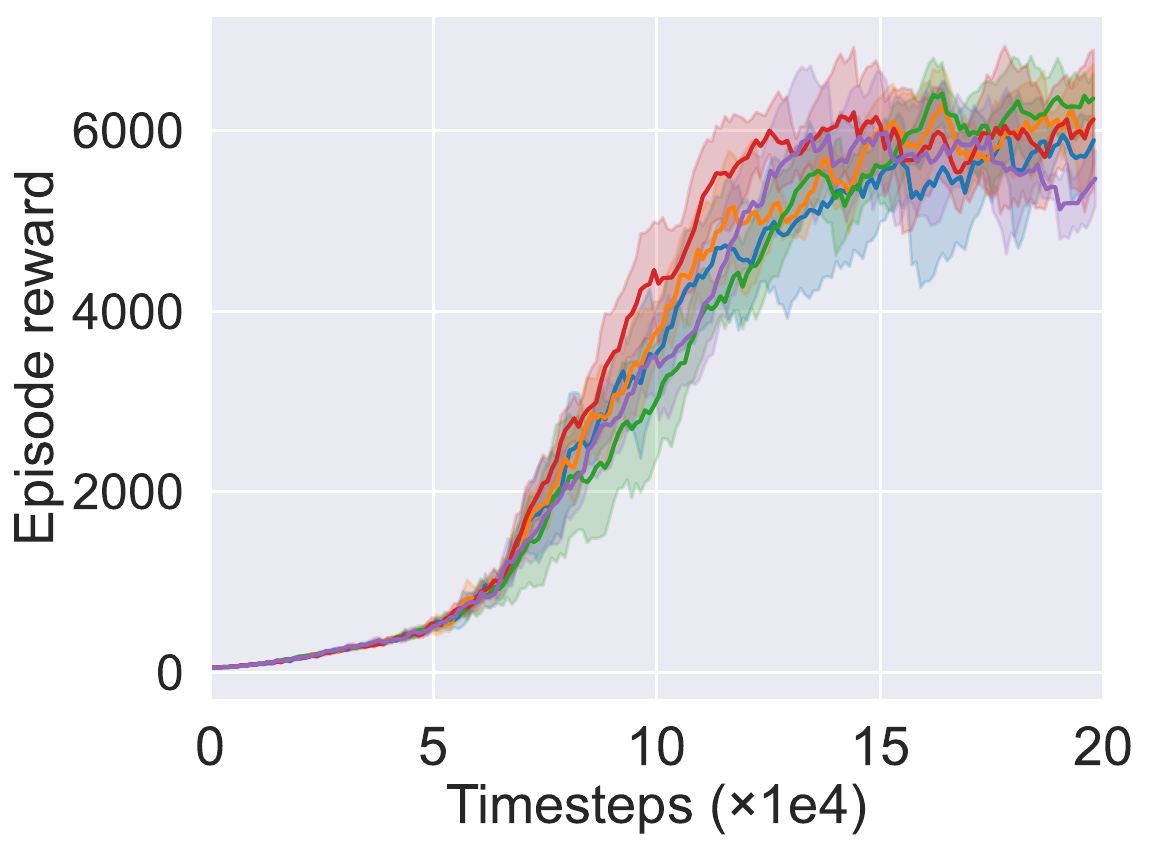}
        \caption{InvertedDoublePendulum}
    \end{subfigure}

    \begin{subfigure}{0.33\textwidth}
        \centering
        \includegraphics[width=\linewidth]{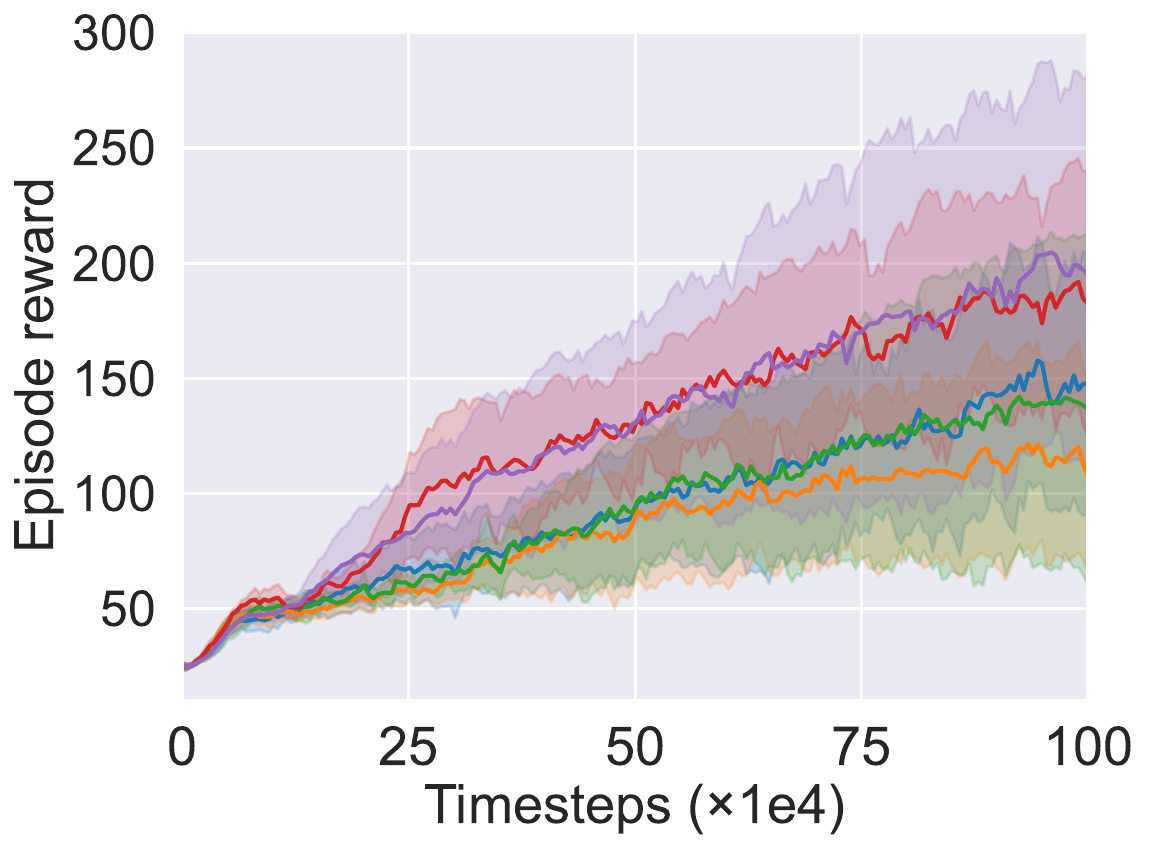}
        \caption{WalkerRun}
    \end{subfigure}
    \hfill
    \begin{subfigure}{0.33\textwidth}
        \centering
        \includegraphics[width=\linewidth]{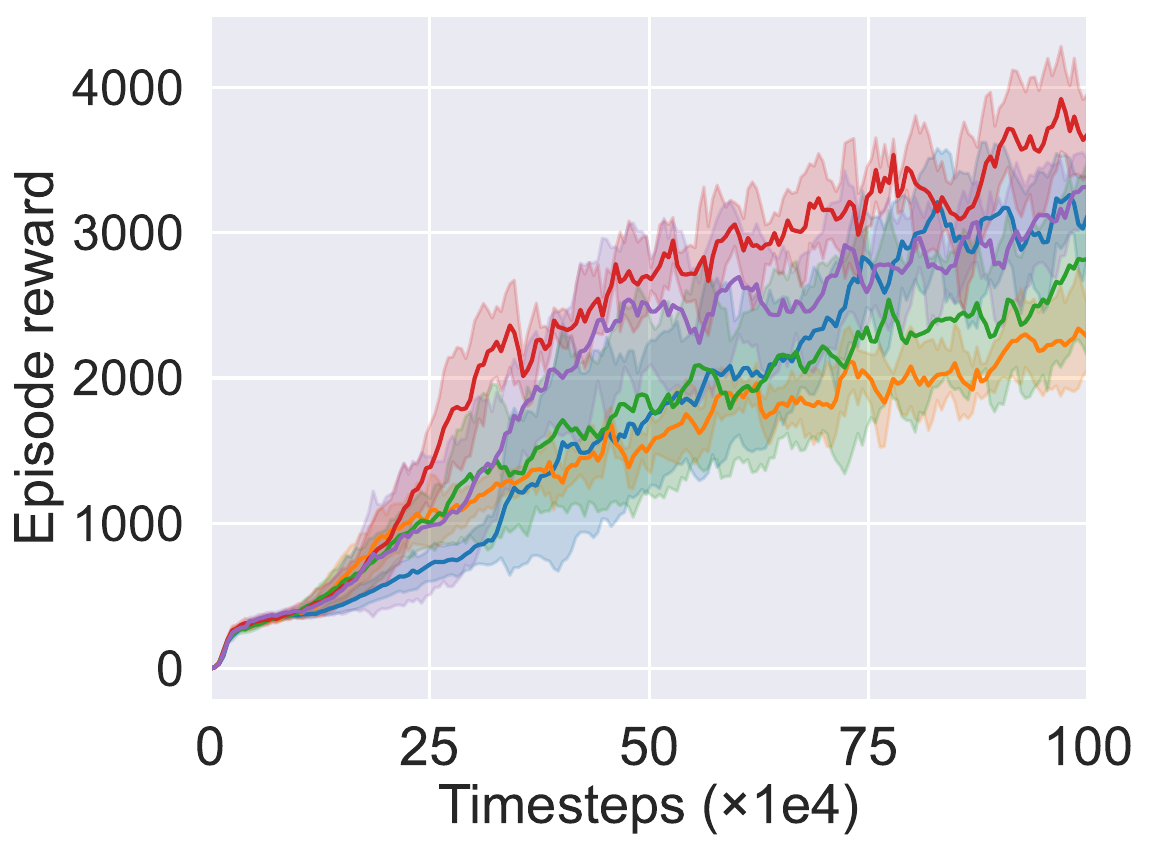}
        \caption{Walker2d}
    \end{subfigure}
    \hfill
    \begin{subfigure}{0.33\textwidth}
        \centering
        \includegraphics[width=\linewidth]{figures/PPO/w/Ant-v4.pdf}
        \caption{Ant}
    \end{subfigure}

    \caption{Sensitive analysis of PPO-based methods on all evaluated tasks.}
    \label{app:w_analysis}
\end{figure}

% ablation
\subsection{Complete Ablation Study Results on Reflex-PPO}
\begin{figure}[!h]
    \centering

    \begin{subfigure}{0.33\textwidth}
        \centering
        \includegraphics[width=\linewidth]{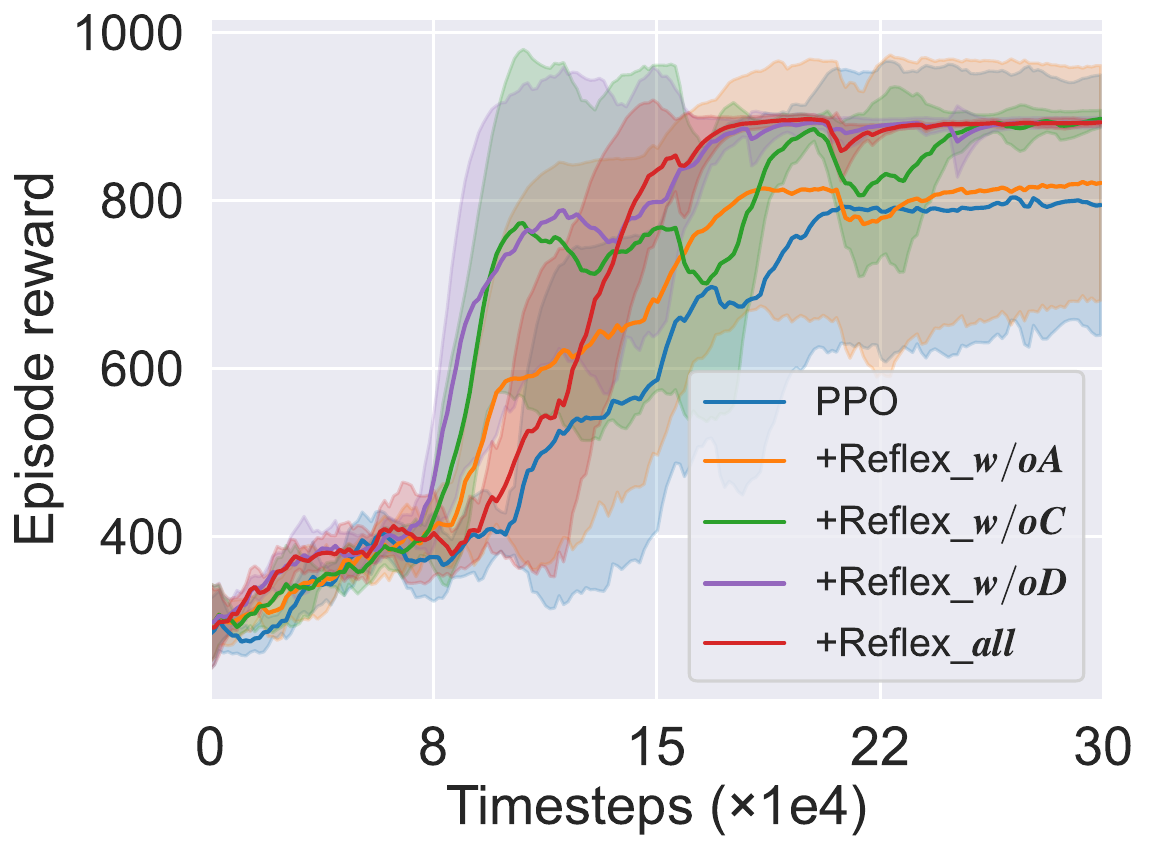}
        \caption{CartpoleBalance}
    \end{subfigure}
    \hfill
    \begin{subfigure}{0.33\textwidth}
        \centering
        \includegraphics[width=\linewidth]{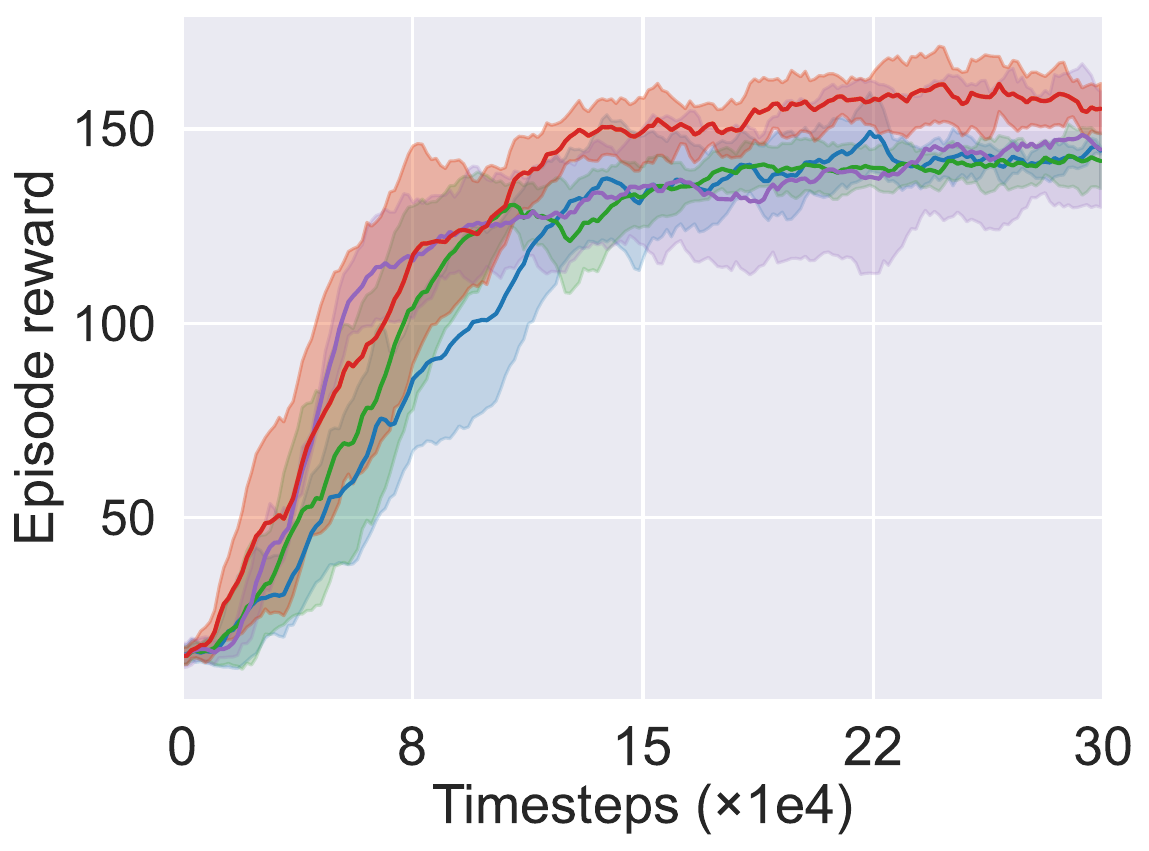}
        \caption{CartpoleThreePoles}
    \end{subfigure}
    \hfill
    \begin{subfigure}{0.33\textwidth}
        \centering
        \includegraphics[width=\linewidth]{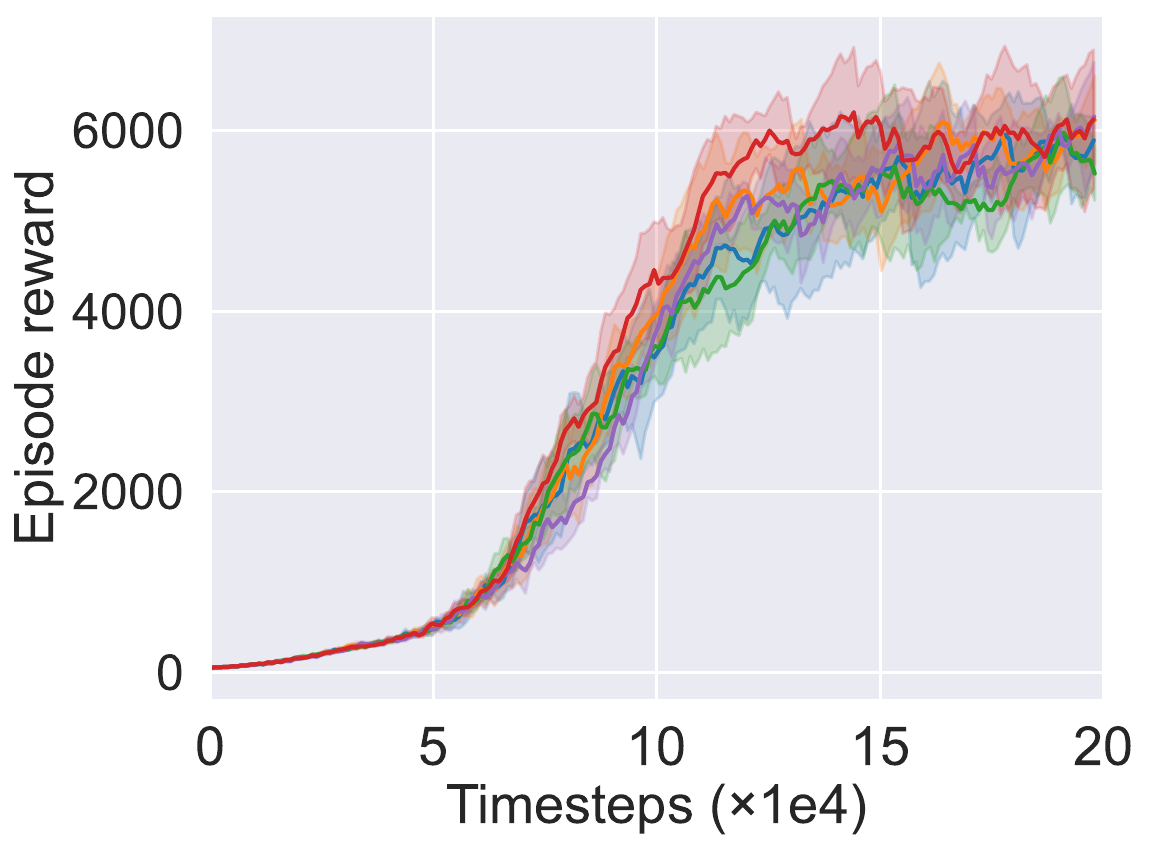}
        \caption{InvertedDoublePendulum}
    \end{subfigure}

    \begin{subfigure}{0.33\textwidth}
        \centering
        \includegraphics[width=\linewidth]{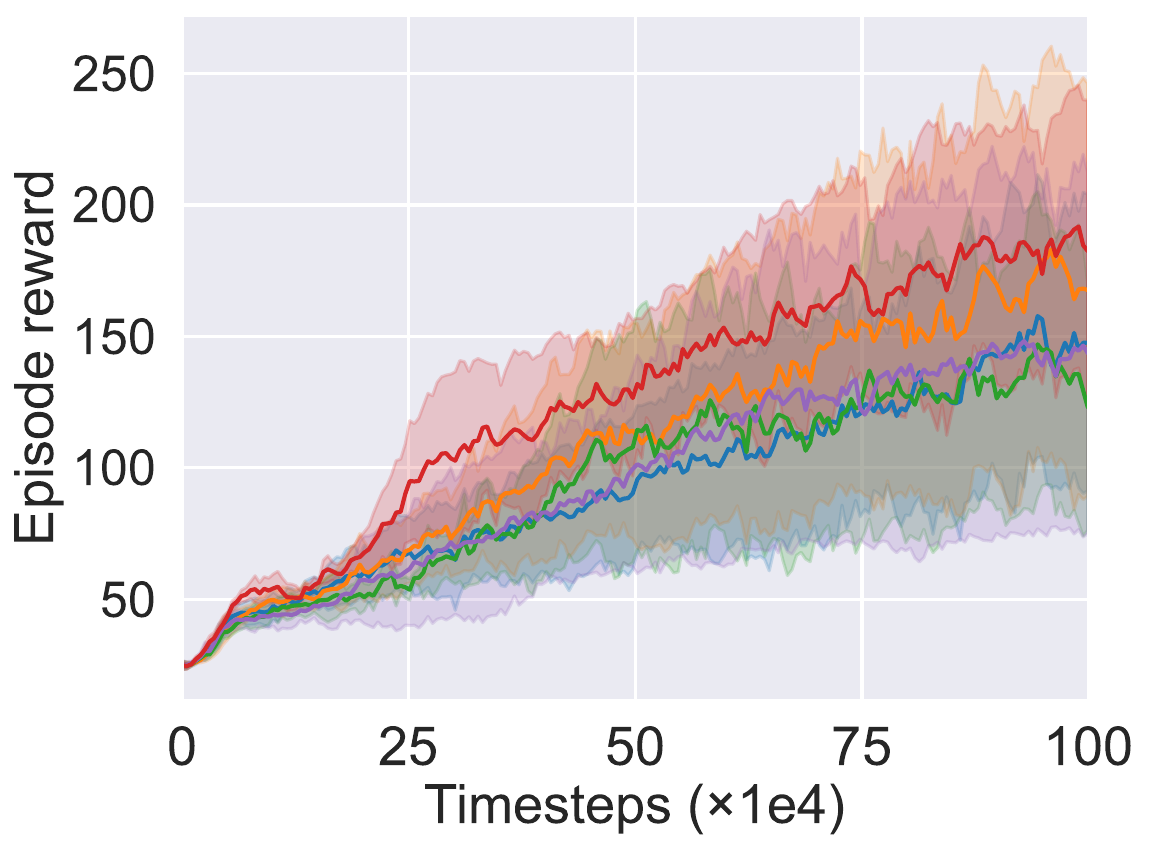}
        \caption{WalkerRun}
    \end{subfigure}
    \hfill
    \begin{subfigure}{0.33\textwidth}
        \centering
        \includegraphics[width=\linewidth]{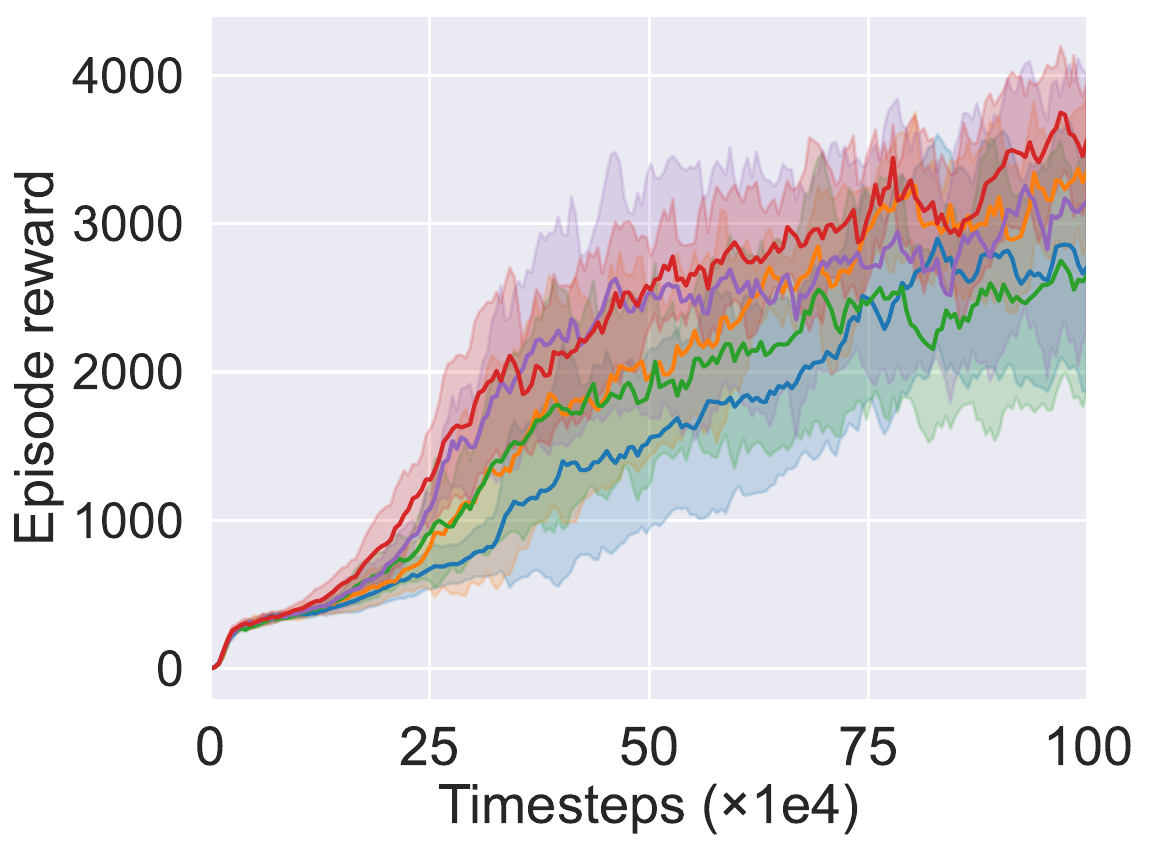}
        \caption{Walker2d}
    \end{subfigure}
    \hfill
    \begin{subfigure}{0.33\textwidth}
        \centering
        \includegraphics[width=\linewidth]{figures/PPO/ablation/Ant-v4.pdf}
        \caption{Ant}
    \end{subfigure}

    \caption{Sensitive analysis of PPO-based methods on all evaluated tasks.}
    \label{app:ablation}
\end{figure}

%%%%%%%%%%%%%%%%%%%%%%%%%%%%%%%%%%%%%%%%%%%%%%%%%%%%%%%%%%%%%%%%%%%%%%%%%%%%%%%
%%%%%%%%%%%%%%%%%%%%%%%%%%%%%%%%%%%%%%%%%%%%%%%%%%%%%%%%%%%%%%%%%%%%%%%%%%%%%%%

\end{document}